%% file: main.tex
\documentclass[10pt,twocolumn,letterpaper]{article}

\usepackage[pagenumbers]{wacv} 
\usepackage{graphicx}
\usepackage{multirow}
\usepackage[inkscapelatex=false]{svg}

\input{preamble}
\definecolor{wacvblue}{rgb}{0.21,0.49,0.74}
\usepackage[pagebackref,breaklinks,colorlinks,allcolors=wacvblue]{hyperref}

\def\wacvPaperID{1411} 
\def\confName{WACV}
\def\confYear{2027}

\title{Real-Time Human Reconstruction and Animation
using Feed-Forward Gaussian Splatting}

\author{Devdoot Chatterjee\\
IIIT Hyderabad, India
\and
Zakaria Laskar\\
IISER Thiruvananthapuram, India
\and
C.V. Jawahar\\
IIIT Hyderabad, India
}

\begin{document}
\maketitle
\input{sec/0_abstract}    
\input{sec/1_intro}

\input{sec/2_related_work}

\input{sec/3_method}
\input{sec/4_experiments}
\input{sec/5_conclusion}
{
    \small
    \bibliographystyle{ieeenat_fullname}
    \bibliography{main}
}
\clearpage

\maketitlesupplementary

\setcounter{section}{0}
\setcounter{figure}{0}
\setcounter{table}{0}
\setcounter{equation}{0}

\renewcommand{\thesection}{S\arabic{section}}
\renewcommand{\thefigure}{S\arabic{figure}}
\renewcommand{\thetable}{S\arabic{table}}
\renewcommand{\theequation}{S\arabic{equation}}

\input{sec/supply}

\end{document}

%% file: preamble.tex
\definecolor{red4}{RGB}{255,199,199}
\definecolor{red3}{RGB}{255,214,214}
\definecolor{red2}{RGB}{255,230,230}
\definecolor{red1}{RGB}{255,242,242}
\definecolor{green1}{RGB}{240,255,240}
\definecolor{green2}{RGB}{225,255,225}
\definecolor{green3}{RGB}{200,255,200}
\definecolor{green4}{RGB}{170,255,170}
\definecolor{cvprblue}{rgb}{0.21,0.49,0.74}

\newcommand{\THtwo}{TH2.1\xspace}%
\newcommand{\AV}{AvRex\xspace}%
\newcommand{\THfour}{TH4.0\xspace}%

\makeatletter
\DeclareRobustCommand\onedot{\futurelet\@let@token\@onedot}
\def\@onedot{\ifx\@let@token.\else.\null\fi\xspace}

\makeatother

\newcommand{\figref}[1]{Fig.~\ref{#1}}

\newcommand{\tabref}[1]{Table~\ref{#1}}

%% file: sec/0_abstract.tex
\begin{abstract}
We present HumanGS, a generalizable feed-forward Gaussian splatting framework for human 3D reconstruction and real-time animation from sparse multi-view RGB images and their associated SMPL-X poses. Unlike prior methods that rely on depth supervision, fixed input views, UV maps, repeated feed-forward inference for each target pose or view, or computationally expensive vertex-to-image cross-attention, HumanGS employs a simple transformer architecture that explicitly associates SMPL-X vertices with multi-scale image features through geometric back-projection. This eliminates the need for large pre-trained human representation models while naturally aggregating complementary information from multiple views. The aggregated vertex features are mapped by a lightweight MLP decoder to a canonical set of 3D Gaussian primitives aligned with SMPL-X vertices. One Gaussian is regularized to remain close to the SMPL-X surface, providing a strong geometric prior and stable correspondence to the parametric body model, while a small set of unconstrained Gaussians per vertex captures geometric details beyond the body surface, such as clothing and hair. The resulting canonical representation is animated efficiently using linear blend skinning and Gaussian rasterization without further network inference. Trained entirely from scratch on only 10K frames from THuman2.1, HumanGS achieves reconstruction quality comparable to or better than state-of-the-art methods while reducing reconstruction time by over 15× compared to recent transformer-based approaches, enabling real-time animation and interactive applications. Code and pre-trained models are
available at: \url{https://github.com/Devdoot57/HumanGS}.

\end{abstract}

%% file: sec/1_intro.tex
\section{Introduction}
\label{sec:intro}

Novel view synthesis (NVS) aims to generate photorealistic renderings of a scene from unseen viewpoints and has seen rapid progress with neural representations such as NeRFs~\cite{mildenhall2021nerf} and 3D Gaussian Splatting (3DGS)~\cite{kerbl20233d}. Extending NVS to humans introduces additional challenges due to articulated motion and complex non-rigid deformations. Beyond reconstructing geometry and appearance from sparse observations, practical systems must support controllable animation under unseen poses while remaining computationally efficient. In this work, we focus on the sparse-view setting, where only a small set of input views of a human—captured under varying poses—is available, reflecting realistic capture scenarios outside controlled multi-view environments.

Existing approaches address this problem through either optimization-based or feed-forward methods. Optimization-based approaches, such as Animatable Gaussians~\cite{li2024animatable}, achieve high-quality animatable human 3D reconstructions using 3DGS but require costly per-subject test-time optimization, limiting scalability. Recent 3D Gaussian Splatting (3DGS) based feed-forward approaches such as GIGA~\cite{giga} Generalizable Human Gaussian (GHG)~\cite{kwon2024generalizable} and GPS-Gaussian~\cite{zheng2023gps} predict Gaussian parameters directly from input images, enabling efficient real-time rendering while avoiding per-subject optimization. However, these methods rely on convolutional architectures, which restrict them to a fixed number of input views and assume a fixed human pose across the inputs, limiting their applicability in realistic capture scenarios where humans naturally appear under varying poses and viewpoints.

Effectively integrating information from multiple views is critical for accurate human reconstruction and novel view synthesis, particularly in sparse-view settings where complementary observations significantly reduce reconstruction ambiguity. Transformer-based Large Reconstruction Models (LRMs) have recently emerged as a powerful paradigm for aggregating information from a variable number of input views. Building on this idea, HumanRAM~\cite{yu2025humanram} incorporates SMPL-X~\cite{pavlakos2019expressive} priors into an LRM for human rendering under novel viewpoints and poses. However, it formulates animation as per-frame image generation, requiring a complete network inference for every target pose and view. More recently, transformer based LHM~\cite{qiu2025lhm} predicts reusable 3D Gaussian representations directly from a single image in a feed-forward pass, achieving state-of-the-art qualitative fidelity. Nevertheless, its single-view formulation is fundamentally limited by incomplete observations, and extending its architecture to multi-view inputs is computationally prohibitive due to dense SMPL-X vertex-to-image cross-attention and reliance on large pre-trained human representation models~\cite{sapiens}. Consequently, there remains a need for a simple transformer-based framework that efficiently leverages sparse multi-view observations while producing an explicit canonical representation that can be animated without repeated network inference.

We propose HumanGS, a simple feed-forward framework that decouples reconstruction from animation via a canonical 3D Gaussian representation. Unlike recent single-image approaches, HumanGS is designed for sparse multi-view inputs, which are readily available in practical capture setups and provide complementary observations for more accurate reconstruction. Given sparse multi-view images and SMPL-X~\cite{pavlakos2019expressive} parameters, our LRM predicts 3DGS parameters aligned with canonical T-pose SMPL-X vertices. Rather than relying on large pre-trained human representation models~\cite{sapiens} and expensive vertex-to-image cross-attention~\cite{qiu2025lhm, Esser2024ScalingRF}, HumanGS exploits the known SMPL-X geometry to explicitly associate vertices with multi-scale image features via back-projection. This simple geometric correspondence naturally aggregates complementary information across views while avoiding the poor scaling of dense cross-attention. The aggregated vertex features are then mapped to 3DGS parameters using a lightweight MLP decoder. Trained entirely from scratch on only 10K frames from THuman2.1, HumanGS reconstructs a reusable canonical asset that is efficiently animated via Linear Blend Skinning and Gaussian rasterization, enabling novel pose and view synthesis without repeated network inference. Experiments on THuman 2.1, 4.0, and AvatarReX show that HumanGS achieves state-of-the-art reconstruction and animation quality while substantially reducing modeling and per-frame synthesis time.

In summary we make the following contributions:
\begin{itemize}
    \item We introduce HumanGS, a simple LRM-based feed-forward framework that reconstructs explicit canonical 3D Gaussians from sparse multi-view images for efficient reconstruction and animation.
    \item We replace costly vertex-to-image cross-attention with explicit SMPL-X-guided feature aggregation via vertex back-projection and a lightweight MLP decoder, enabling training efficiently.
    \item We decouple reconstruction from rendering, enabling real-time skinning-based animation without repeated network passes.
    \item Despite training on only 10K frames from THuman2.1, HumanGS leverages complementary multi-view observations to outperform single-view LHM~\cite{qiu2025lhm} while reducing reconstruction time by over $15\times$ (0.29\,s vs.\ 4.59\,s).
\end{itemize}

%% file: sec/2_related_work.tex
\section{Related Work}
\label{sec:related_work}

\noindent \textbf{Human 3D Reconstruction.}
Human 3D reconstruction has been widely studied for applications such as virtual avatars, telepresence, and digital content creation. Early methods relied on dense multi-view capture systems in controlled environments to reconstruct high-fidelity geometry and appearance~\cite{collet2015high,guo2019relightables}. While these approaches achieve highly accurate reconstructions, they require specialized hardware and dense camera setups, which limits scalability and practical deployment.

Recent advances in Neural Radiance Fields (NeRF)~\cite{mildenhall2021nerf} have inspired extensive work~\cite{HNeRF:NeurIPS:2021,NARF:ICCV:2021,ANeRF:NeurIPS:2021,jiang2022neuman,peng2021animatable,ARAH:ECCV:2022,guo2023vid2avatar,li2022tava,peng2022animatable,jiang2022instantavatar,weng2022humannerf,yu2023monohuman} on human reconstruction, often leveraging parametric priors like SMPL~\cite{zhang2021editable, peng2021neural, peng2021animatable, liu2021neuralactor, weng2022humannerf, weng2020vid2actor} to model articulation. Despite strong quality, NeRF's expensive optimization and slow rendering prompted exploration into faster representations like Instant-NGP~\cite{mueller2022instant,jiang2022instantavatar,instant_nvr}. More recently, 3D Gaussian Splatting (3DGS)~\cite{kerbl20233d} has emerged as a highly efficient alternative. Multiple methods extend 3DGS to human reconstruction~\cite{kocabas2024hugs, moreau2024human, Pang_2024_CVPR, hu2024gauhuman, li2024gaussianbody, qian20233dgsavatar, shao2024splattingavatar, li2024animatable}, using SMPL priors to regularize geometry and support pose-driven deformation. Notably, Animatable Gaussians~\cite{li2024animatable} achieve high-fidelity pose-dependent animation. However, while offering real-time rendering, these optimization-based 3DGS pipelines still require minutes to hours of per-subject modeling time, strictly limiting practical scalability.

To address this limitation, recent works explore generalizable feed-forward reconstruction models that directly predict scene representations in a single forward pass, reducing modeling time to seconds. Our proposed HumanGS follows this direction and enables fast reconstruction and real-time animation through an explicit canonical Gaussian representation.

\noindent \textbf{Generalizable Feed-Forward Human Reconstruction.}
 Early approaches learn image-conditioned radiance fields using pixel-aligned features~\cite{saito2019pifu,yu2021pixelnerf,wang2022attention} or cost volumes~\cite{chen2021mvsnerf,wang2024learning} extracted from multi-view observations. These ideas have been extended to human reconstruction~\cite{kwon2021neural,zhao2022humannerf,cheng2022generalizable,chen2022geometry,gao2022mps,kwon2023neural,chen2023gm,gao2023neural}, enabling generalization across subjects without per-subject training. Notably, these image-conditioned radiance fields use SMPL as a geometric prior for robust aggregation of multi-view image features. More recently, transformer-based Large Reconstruction Models (LRMs)~\cite{hong2023lrm} have demonstrated strong performance in generalizable novel view synthesis. LVSM~\cite{jin2024lvsm} demonstrates the use of LRM for novel view synthesis. HumanRAM~\cite{yu2025humanram} adapts LRM-based LVSM architecture for human reconstruction and animation by also conditioning on SMPL parameters, enabling synthesis under novel poses and viewpoints. While the above methods, especially HumanRAM achieves good reconstruction quality, animation is formulated as per-frame image generation, requiring repeated network inference for each target view and pose.

Several feed-forward approaches instead predict 3D Gaussian representations directly from images. These methods regress pixel-aligned~\cite{zheng2023gps} or UV map-aligned~\cite{ghg, giga} 3DGS parameters. However, these existing methods are based on convolutional architectures and therefore tied to a fixed number of input views, while also assuming fixed-pose inputs. Concurrently, the Large Animatable Human Reconstruction Model (LHM)~\cite{qiu2025lhm} achieves state-of-the-art qualitative fidelity by leveraging a multimodal transformer to fuse geometric and image tokens for direct prediction of 3D Gaussians. However, its dense cross-attention between $\sim$40K canonical SMPL-X vertices and image patch features is computationally expensive and scales poorly to multi-view inputs. A similar architecture was independently explored in the concurrent work LCA~\cite{li2026lca}. Their results (Table 2 in~\cite{li2026lca}) show that multi-view LHM consistently outperforms its single-view counterpart, highlighting the benefit of integrating multiple observations. However, naively extending LHM to multiple views by concatenating image tokens is prohibitively expensive due to its dense vertex-to-image cross-attention, making such experiments infeasible on our hardware. Furthermore, as LCA has not released its training or evaluation code, a fair quantitative comparison with HumanGS is currently not possible.

Instead of dense vertex-image cross-attention, we explicitly establish vertex-image correspondences by back-projecting SMPL-X vertices onto multi-scale image feature maps, eliminating costly cross-attention while naturally supporting sparse multi-view reconstruction. HumanGS directly predicts canonical (T-pose) SMPL-vertex-aligned 3D Gaussians, decoupling reconstruction from animation and enabling efficient multi-view reconstruction and real-time animation via LBS and Gaussian rasterization without repeated network inference.

%% file: sec/3_method.tex
\section{Method}

\begin{figure*}[t]
\centering
\includegraphics[width=\textwidth]{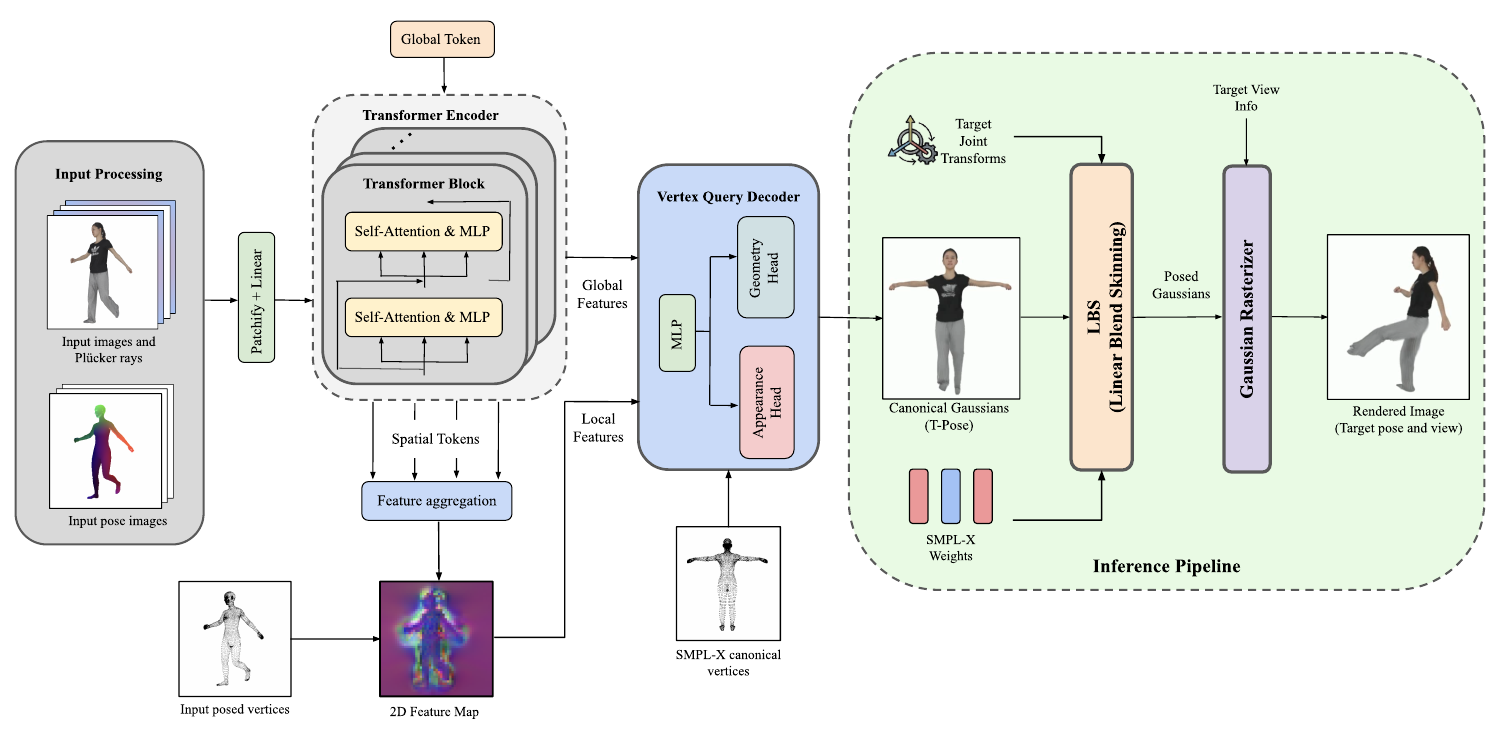}
\caption{\textbf{Overview of the HumanGS architecture and explicit inference pipeline.} A Transformer Encoder processes sparse input images and pose maps to extract global 2D feature maps. Canonical SMPL-X~\cite{pavlakos2019expressive} vertices project onto these maps to sample local geometry-aware features. A Vertex Query Decoder predicts 3D Gaussian attributes in the canonical T-pose space. During inference (green dashed box), these Gaussians are animated via Linear Blend Skinning (LBS) with target pose transforms and rasterized, bypassing neural network evaluation.}
\vspace{-5mm}
\label{fig:architecture}
\end{figure*}

We propose HumanGS, a unified feed-forward framework for high-fidelity 3D human reconstruction and real-time animation (see Figure~\ref{fig:architecture}). Unlike previous state-of-the-art methods like HumanRAM~\cite{yu2025humanram} which formulate animation as a 2D image-to-image translation task conditioned on 3D priors, our approach explicitly predicts a canonical 3D Gaussian Splatting (3DGS) asset. This asset is intrinsically animatable via Linear Blend Skinning (LBS), decoupling the heavy feature extraction process from the rendering loop and enabling real-time performance.

\subsection{Background}

Our method builds upon recent advancements in Large Reconstruction Models and neural texture-based pose conditioning.

\noindent \textbf{Large View Synthesis Model (LVSM)}. The foundation of our architecture is the Large View Synthesis Model (LVSM), a scalable transformer-based framework designed for generalizable 3D reconstruction. LVSM operates by tokenizing input multi-view images and their corresponding camera parameters into a latent sequence.
Formally, given $N$ input images $\mathbf{I} \in \mathbb{R}^{N \times H \times W \times 3}$ and cameras parameterized as Plücker ray embeddings $\mathbf{P} \in \mathbb{R}^{N \times H \times W \times 6}$~\cite{plucker}, LVSM projects $p \times p$ patches into input tokens $\mathbf{x}_{i,j} \in \mathbb{R}^D$:
\begin{equation}
\mathbf{x}_{i,j} = \text{Linear}_{\text{inp}}([\mathbf{I}_{i,j}, \mathbf{P}_{i,j}])
\end{equation}
where $[\cdot]$ denotes concatenation~\cite{vit}. These tokens are processed by a transformer encoder to learn a 3D-aware global representation. While the original LVSM targets novel view synthesis via a pixel-decoder, we leverage its powerful encoder to extract robust 3D features from sparse inputs.

\noindent\textbf{SMPL-X and Linear Blend Skinning (LBS).} To introduce structural priors, we utilize the SMPL-X parametric model. A human body is defined by shape $\beta$ and pose $\theta$ parameters, yielding a mesh of $N_{v} = 10,475$ vertices. Animation is achieved via Linear Blend Skinning (LBS)~\cite{lbs}. The posed position of a canonical T-pose vertex $v_{cano}$ is computed as:
\begin{equation}
\mathbf{v}_{posed} = \sum_{j=1}^{J} w_{j} \mathbf{G}_j(\theta) \mathbf{v}_{cano}
\end{equation}
where $w_j$ are the skinning weights associated with joint $j$, and $\mathbf{G}_j(\theta)$ is the world transformation matrix of joint $j$. This differentiable transformation allows us to explicitly bind generated 3D primitives to the parametric surface.

\noindent\textbf{Pose Image Map.}
Establishing 2D-3D correspondence is a critical challenge in feed-forward human reconstruction. We adopt HumanRAM's \textit{Neural Texture} and \textit{Pose Image} mechanism. We define a learnable neural texture $\mathcal{T}$ via three orthogonal planes $\{\mathcal{T}_{xy}, \mathcal{T}_{xz}, \mathcal{T}_{yz}\}$ in the canonical SMPL-X space. For any point $\mathbf{v}$ on the canonical surface, a texture feature $F(\mathbf{v})$ is sampled via bilinear interpolation:
\begin{equation}
F(\mathbf{v}; \mathcal{T}) = [\mathcal{T}_{xy}(\mathbf{v}_{xy}), \mathcal{T}_{xz}(\mathbf{v}_{xz}), \mathcal{T}_{yz}(\mathbf{v}_{yz})]
\end{equation}
To encode the subject's pose, we rasterize these features onto the input viewpoints~\cite{thies2019deferred}. Rendering the registered SMPL-X mesh produces a correspondence map used to sample $\mathcal{T}$, yielding a Pose Image Map $\mathbf{F}_{pose} \in \mathbb{R}^{H \times W \times C}$. These maps align the canonical body's semantic identity with the input pixel space, guiding the network to disentangle appearance from pose.

\noindent\textbf{Canonical T-pose \cite{pavlakos2019expressive}.}
Unlike image-space methods that reconstruct directly in the target view, we define our prediction target in a shared \textit{Canonical T-pose} space. This space corresponds to the zero-pose configuration of the SMPL-X model ($\theta = \mathbf{0}$). By predicting geometry in this neutral frame, we ensure that the generated 3D asset is disentangled from the specific pose observed in the input images, facilitating consistent animation across novel poses.

\subsection{HumanGS}

HumanGS reformulates the reconstruction task: instead of regressing pixels for a target view (as in HumanRAM), we regress a set of 3D Gaussian primitives in the \textit{canonical space}. This allows the generated asset to be re-posed and rendered from any viewpoint without re-evaluating the transformer. Our architecture consists of three stages: feature encoding, vertex-aligned sampling, and canonical decoding.

\noindent\textbf{Transformer Encoder.}
Similar to HumanRAM, we concatenate the RGB image, Plücker rays, and the rasterized Pose Image Maps along the channel dimension. These are patchified and fed into a transformer encoder. The encoder outputs a sequence of spatial tokens $\mathbf{Z}_{spatial}$, which we reshape into a 2D feature map $\mathbf{S} \in \mathbb{R}^{N_{views} \times H' \times W' \times D}$ representing the global context of the subject.

\begin{figure}[t]
\centering
\includegraphics[width=\linewidth]{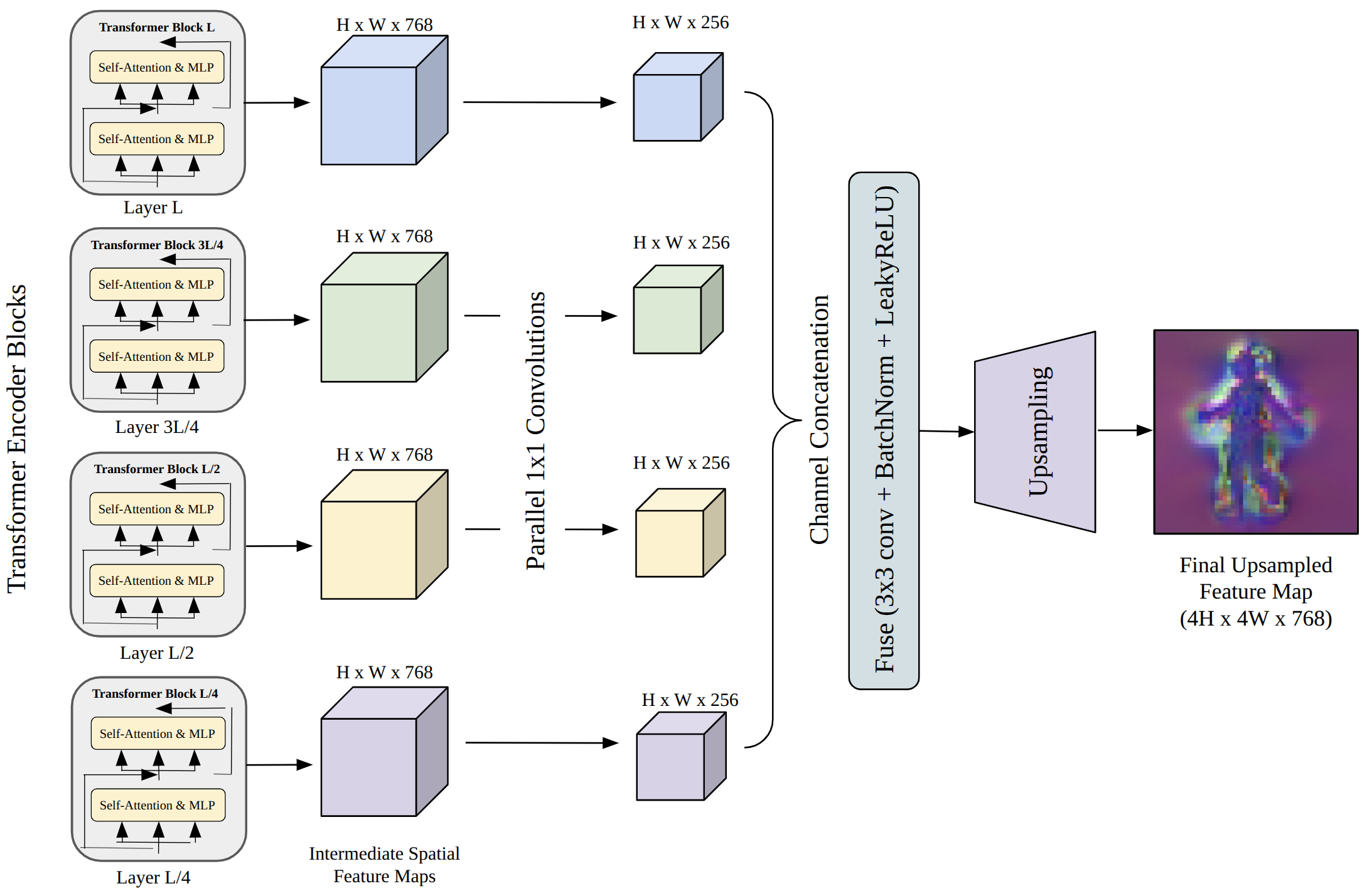}
\caption{\textbf{Feature Aggregation.} Tokens from four uniform transformer depths are reshaped into 2D maps, projected via parallel $1 \times 1$ convolutions, concatenated, and fused. Progressive upsampling quadruples the resolution to produce a dense feature map for vertex-aligned sampling.}
\vspace{-5mm}
\label{fig:intermediate_features}
\end{figure}


\noindent \textbf{Feature Aggregation Module.} As depicted in Figure~\ref{fig:architecture} and detailed in Figure~\ref{fig:intermediate_features}, we aggregate spatial tokens from multiple network depths to prevent the over-smoothing of high-frequency textural details common in standard vision transformers. From an $L$-layer encoder, 1D tokens extracted at layers $L/4$, $L/2$, $3L/4$, and $L$ are reshaped into 2D spatial maps. Each map is projected to 256 channels via parallel $1 \times 1$ convolutions, concatenated, and fused using a $3 \times 3$ convolution, batch normalization, and LeakyReLU activation. To facilitate vertex-aligned sampling, this fused representation undergoes two upsampling stages---each comprising a $4 \times 4$ transposed convolution (stride $2$, padding $1$) and a refining $3 \times 3$ convolution. This quadruples the spatial resolution (e.g., $64 \times 64$ to $256 \times 256$) to output a dense 768-channel feature map, onto which canonical SMPL-X vertices are projected to reliably sample local geometry-aware features.

\noindent \textbf{Vertex-Aligned Feature Sampling.}
To bridge the gap between image space and the 3D surface, we query features at the locations of the SMPL-X vertices. For each vertex $\mathbf{v}_i$ in the \textit{canonical} mesh, we compute its projection onto the feature map $\mathbf{S}$ using the camera parameters of the input views. We obtain a local feature vector $\mathbf{f}_{local}^{(i)}$ by sampling $\mathbf{S}$ at the projected coordinates and pooling across visible views. This provides a dense, geometry-aware feature representation for every vertex on the parametric model.

\noindent\textbf{Vertex Query Decoder}
The core innovation of HumanGS is the Decoder $\Phi$. It is designed as a multi-head Multi-Layer Perceptron (MLP) that predicts the attributes of 3D Gaussians associated with each SMPL-X vertex.
For each vertex $\mathbf{v}_i$, the decoder first aggregates the global context vector (from the Global token), the local vertex feature $\mathbf{f}_{local}^{(i)}$, and the canonical vertex position $\mathbf{v}_{i}$ via an MLP. This fused representation is then fed into two parallel branches:

\begin{enumerate}
    \item \textbf{Geometry Head:} Predicts the spatial attributes for $K$ Gaussian primitives per vertex:
    \begin{equation}
    \{ (\Delta \mu_{i,k}, \mathbf{q}_{i,k}, \mathbf{s}_{i,k}) \}_{k=1}^K = \Phi_{geo}(\mathbf{h}_i)
    \end{equation}
    where $\Delta \mu_{i}$ is the positional offset with respect to the canonical vertex, $\mathbf{v}_{i}$, $\mathbf{q}$ is the rotation quaternion, and $\mathbf{s}$ is the scaling factor.
    
    \item \textbf{Appearance Head:} Predicts the visual attributes:
    \begin{equation}
    \{ (\alpha_{i,k}, \mathbf{c}_{i,k}) \}_{k=1}^K = \Phi_{app}(\mathbf{h}_i)
    \end{equation}
    where $\alpha$ is opacity and $\mathbf{c} \in [0, 1]^3$ is the view-independent RGB color.
\end{enumerate}

The absolute canonical position of the $k$-th Gaussian attached to vertex $i$ is defined as:
\begin{equation}
\mu_{i,k}^{cano} = \mathbf{v}_{i} + \Delta \mu_{i,k}
\end{equation}
We define two types of Gaussians per vertex to capture both the body surface and loose clothing details:
\begin{enumerate}
    \item \textbf{Tight Gaussians:} Highly regularized to remain close to the parent vertex ($\Delta \mu \approx 0$), preserving the underlying body shape.
    \item \textbf{Free Gaussians:} Allowed larger offsets to model non-rigid deformations like hair, skirts, or jackets that deviate from the SMPL-X topology.
\end{enumerate}

In all our experiments, we empirically set the total number of Gaussians per vertex to $K = 5$. To effectively balance surface fidelity and geometric flexibility, the first primitive ($K = 1$) is strictly enforced as a tight Gaussian, while the remaining four primitives are optimized as free Gaussians.


\noindent \textbf{Real-time Animation and Rendering.}
Once the canonical Gaussians $\{\mathcal{G}^{cano}\}$ are predicted, animation is strictly explicit (see the Inference Pipeline in Figure~\ref{fig:architecture}). To render the character in a novel target pose $\theta_{tgt}$, we apply LBS to transform the canonical means and rotations into world space. Crucially, each Gaussian inherits the skinning weights $\mathbf{w}_i$ of its parent vertex $\mathbf{v}_i$:
\begin{equation}
\mu_{i,k}^{posed} = \text{LBS}(\mu_{i,k}^{cano}, \theta_{tgt}, \mathbf{w}_i)
\end{equation}
\begin{equation}
\mathbf{q}_{i,k}^{posed} = \text{LBS}_{rot}(\mathbf{q}_{i,k}^{cano}, \theta_{tgt}, \mathbf{w}_i)
\end{equation}
The posed Gaussians are then rasterized using differentiable 3D Gaussian Splatting to produce the final image $\hat{\mathbf{I}}$. This pipeline allows for rendering at high frame rates simply by updating the transformation matrices $\mathbf{G}(\theta)$, completely bypassing the neural network during inference.

\noindent \textbf{Loss Function}
We train HumanGS end-to-end using a combination of reconstruction and regularization losses.
Following HumanRAM, we employ a weighted combination of Mean Squared Error (MSE) and Perceptual Loss to supervise the image reconstruction quality:
\begin{equation}
\mathcal{L}_{recon} = \frac{1}{M} \sum_{i=1}^{M} \left( \mathcal{L}_{\text{MSE}}(\hat{\mathbf{I}}_i, \mathbf{I}_i) + \lambda_{\text{perc}} \cdot \mathcal{L}_{\text{Perc}}(\hat{\mathbf{I}}_i, \mathbf{I}_i) \right)
\end{equation}
where $M$ is the number of target views, and $\mathcal{L}_{\text{Perc}}$ computes the $L_1$ distance between features extracted from a pre-trained VGG-19 network. In our experiments, we set $\lambda_{\text{perc}} = 1.0$.

To prevent "floating" artifacts common in sparse-view optimization, we adopt the \textit{Tightness Regularization} proposed in Gaussian Splatting Transformer (GST)~\cite{gst}. This term penalizes large deviations of the "tight" Gaussians from their parent vertices, ensuring the geometry respects the SMPL-X topology where appropriate:
\begin{equation}
\mathcal{L}_{reg} = \frac{1}{N_v} \sum_{i=1}^{N_v} \| \Delta \mu_{i, 1} \|_2^2
\end{equation}
The total loss is $\mathcal{L} = \mathcal{L}_{recon} + \lambda_{reg} \mathcal{L}_{reg}$, where we set $\lambda_{reg} = 0.1$.

%% file: sec/4_experiments.tex
\section{Experiments}

\subsection{State-of-the-art}
We compare against seven state-of-the-art baselines: \textbf{i) LVSM}~\cite{jin2024lvsm}, a transformer-based LRM for static-pose novel-view synthesis; \textbf{ii) HumanRAM}~\cite{yu2025humanram}, an SMPL-conditioned LRM for reconstruction and animation; \textbf{iii) GHG}~\cite{ghg}, a CNN predicting 3D Gaussians from fixed views for static reconstruction; \textbf{iv) GST}~\cite{gst}, a single-image transformer predicting SMPL-aligned Gaussians; \textbf{v) GIGA}~\cite{giga}, a recent feed-forward 3D Gaussian approach for human reconstruction; \textbf{vi) Animatable Gaussians}~\cite{li2024animatable}, an optimization baseline animating canonical Gaussians via LBS; and \textbf{vii) LHM}~\cite{qiu2025lhm}, a state-of-the-art single-image feed-forward model for rapid human animation.


\subsection{Datasets and Metrics}

We evaluate our method on three datasets: \textbf{i) THuman 2.1} (\THtwo)~\cite{thuman21}, a static-pose reconstruction dataset comprising 60 views of 2,445 subjects (split into 2,300 for training and 145 for testing); \textbf{ii) THuman 4.0} (\THfour)~\cite{thuman4}, which supports both novel-view and novel-pose evaluation, providing 24 views of 3 test subjects across more than 2,000 poses; and \textbf{iii) AvatarReX} (\AV)~\cite{avatarrex}, a dynamic dataset featuring 16 views of 4 subjects in over 1,600 poses, were also used for testing.



\begin{figure*}[t]
    \centering
    \begin{subfigure}{0.8\textwidth}
        \centering
        \includegraphics[width=\linewidth]{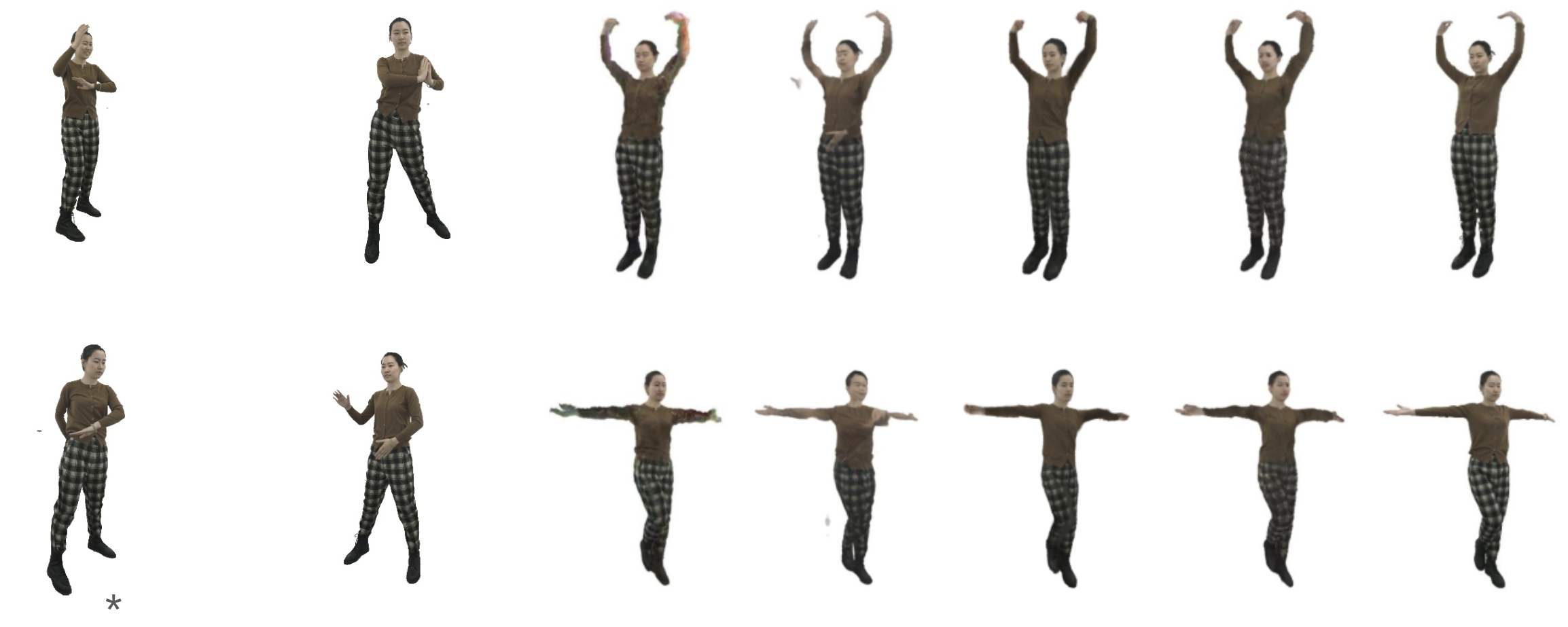}
    \end{subfigure}
    \\ \vspace{0.18cm}
    \begin{subfigure}{0.8\textwidth}
        \centering
        \includegraphics[width=\linewidth]{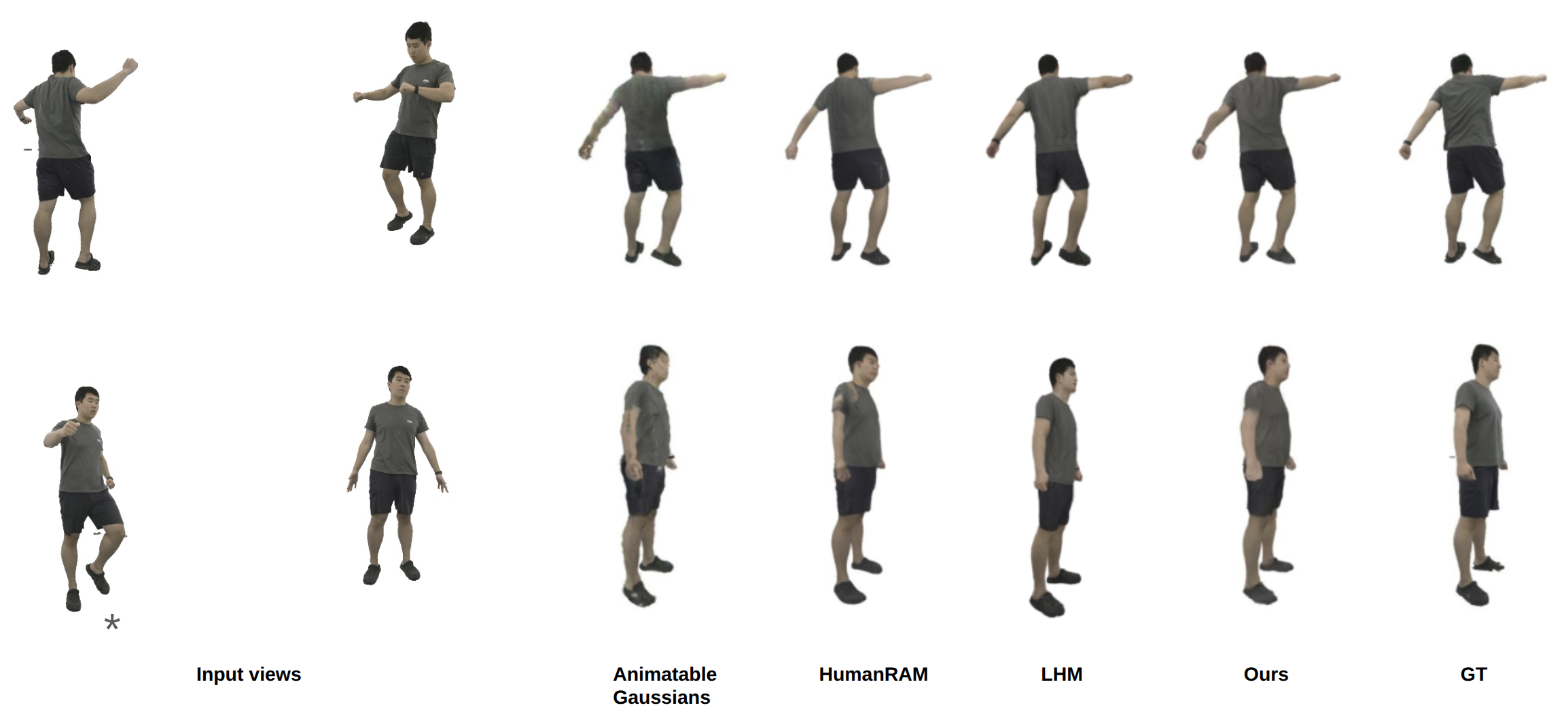}
    \end{subfigure}

    \caption{\textbf{Qualitative Generalization on AvatarReX.} We visualize the animation capabilities on unseen subjects. \textbf{Left:} Input reference view ($^*$denotes the single input for LHM; other methods use 4 views). \textbf{Right:} Re-enactment results driven by a novel motion sequence.}
    \label{fig:avatarex_qualitative}
\end{figure*}

\subsection{Cross-Dataset Generalization and Animation Performance}

\begin{figure*}[t]
    \centering
    \begin{subfigure}{0.8\textwidth}
        \centering
        \includegraphics[width=\linewidth]{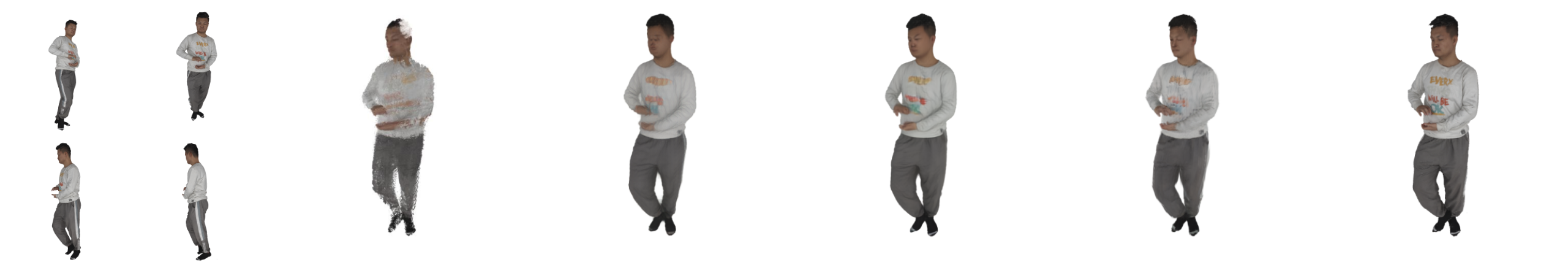}
    \end{subfigure}
    \\ \vspace{0.18cm} 
    \begin{subfigure}{0.8\textwidth}
        \centering
        \includegraphics[width=\linewidth]{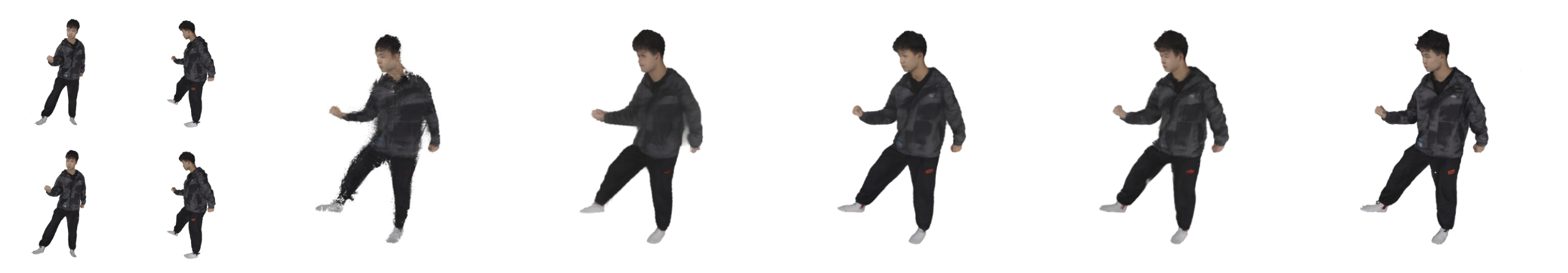}
    \end{subfigure}
    \\ \vspace{0.18cm}
    \begin{subfigure}{0.8\textwidth}
        \centering
        \includegraphics[width=\linewidth]{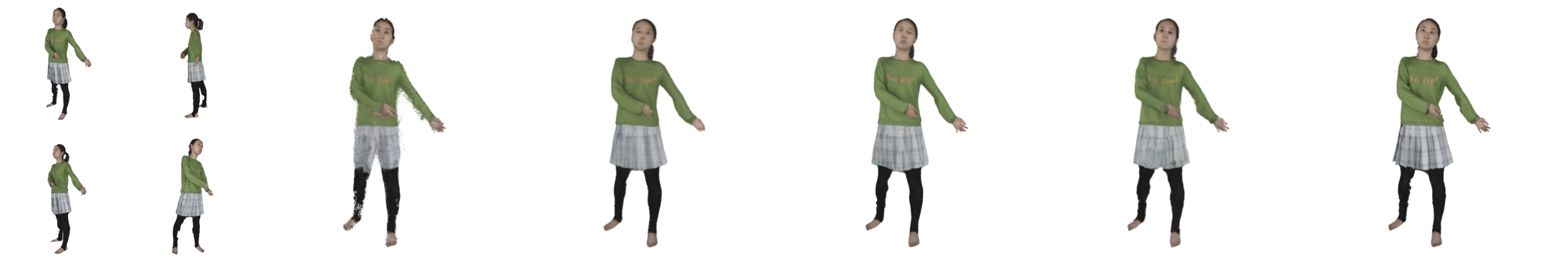}
    \end{subfigure}
    \\ \vspace{0.18cm}
    \begin{subfigure}{0.8\textwidth}
        \centering
        \includegraphics[width=\linewidth]{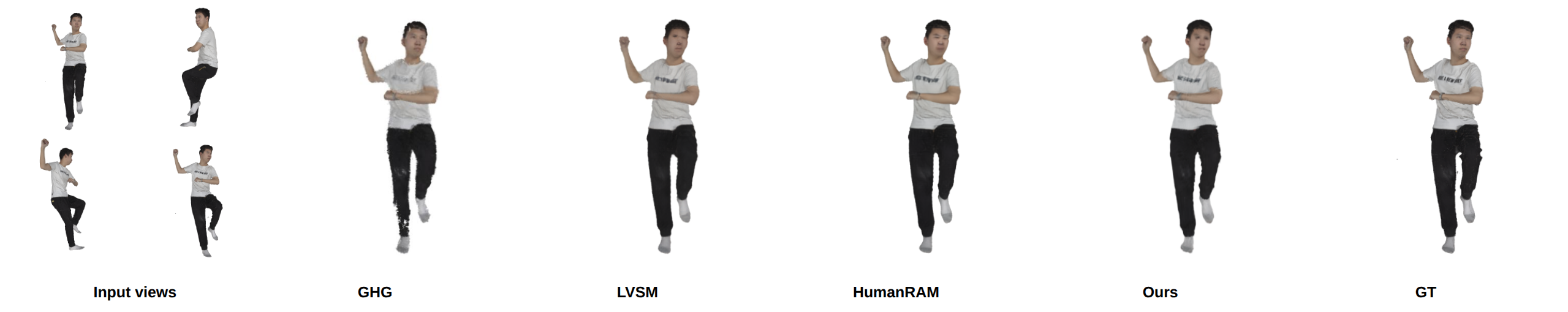}
    \end{subfigure}
    \caption{\textbf{Qualitative comparison on THuman2.1} For each subject, the left block shows the sparse input views. The right block displays novel view synthesis results from different methods.}
    \label{fig:qualitative_comparison}
\end{figure*}

\noindent \textbf{Evaluation on \AV.} We first evaluate cross-dataset generalization on \AV~(\figref{fig:avatarex_qualitative}), testing realistic human motion sequences where subjects appear in diverse poses and viewpoints. Except for the pre-trained LHM---which is provided a front-facing input to ensure a fair comparison against 4-view methods---all feed-forward models are trained solely on THuman 2.1 and evaluated without fine-tuning. We focus on the \textit{Animation} setting---predicting unseen poses and viewpoints from multi-view inputs of varying poses---which jointly evaluates reconstruction and animation capabilities under real-world conditions. This protocol strictly evaluates methods supporting varying-pose inputs and novel-pose synthesis (HumanRAM, Animatable Gaussians, LHM, and HumanGS), naturally excluding static-pose baselines like LVSM and GHG.

\begin{table}[h]
\centering
\resizebox{\columnwidth}{!}{
\begin{tabular}{l|ccc|cc}
\hline
 & \multicolumn{5}{c}{\THtwo $\rightarrow$ \AV} \\
\cline{2-6}
Methods
 & PSNR $\uparrow$ 
 & SSIM $\uparrow$ 
 & LPIPS $\downarrow$
 & Modeling (s) $\downarrow$ 
 & Synthesis (s) $\downarrow$ \\
\hline

LHM* & 23.72 & 0.95 & 0.045 & 4.59 & 0.053 \\
HumanRAM & 24.58 & 0.96 & 0.042 & - & 0.287 \\
Animatable Gaussians & 25.51 & 0.96 & 0.041 & 12000 & 0.245 \\
HumanGS  & 25.24 & 0.96 & 0.04 & 0.29 & 0.048 \\

\hline
\end{tabular}
}
\caption{\textbf{Quantitative evaluation on \THtwo $\rightarrow$ \AV.} HumanGS demonstrates robust cross-dataset generalization, matching the rendering fidelity of optimization-based methods while drastically reducing computational time. (*Note: LHM was evaluated using its official pre-trained model, which leverages a single-image input and was trained on a large-scale dataset of $\sim$300k videos.)}
\vspace{-5mm}
\label{tab:animation_av}
\end{table}

\noindent \textbf{Performance and Efficiency.}
Alongside quality metrics, we evaluate modeling and synthesis times. Modeling time is the duration to construct a canonical 3D representation (applicable to Animatable Gaussians, LHM, and HumanGS), while synthesis time is the per-frame rendering duration. For HumanGS and Animatable Gaussians, synthesis requires only LBS and rasterization, whereas HumanRAM requires full network inference per frame.

As shown in~\tabref{tab:animation_av} and~\figref{fig:avatarex_qualitative}, HumanGS achieves synthesis quality comparable to HumanRAM, Animatable Gaussians, and LHM, but with significantly greater efficiency. Though LHM operates on a challenging single-image input, our 4-view model yields slightly better metrics. Modeling a subject takes Animatable Gaussians ${\sim}12,000$s (3-4 hours) due to per-subject optimization. In contrast, HumanGS predicts a canonical model in just 0.29s---nearly $15\times$ faster than LHM's 4.59s. During inference, Animatable Gaussians and HumanRAM need ${\sim}0.3$s per frame, whereas HumanGS requires only 0.048s, remaining highly competitive with LHM's 0.053s.

\noindent \textbf{Evaluation on \THfour.}
We observe consistent trends on \THfour as shown in~\tabref{tab:quantitative_results-th4}. Although both HumanGS and HumanRAM are trained solely on \THtwo - which contains only reconstruction data (novel views under fixed pose) - they generalize well to \THfour, which includes both novel views and poses under the \textit{Animation} evaluation setup. HumanGS maintains comparable animation quality to HumanRAM while achieving substantially faster synthesis. These results demonstrate that our approach combines strong cross-dataset generalization with significantly improved computational efficiency in synthesizing animations.

\begin{table*}[t]
    \centering
    \begin{minipage}[b]{0.48\textwidth}
        \centering
        \resizebox{\linewidth}{!}{
        \begin{tabular}{lcccccc}
            \toprule
            & \multicolumn{3}{c}{Reconstruction} & \multicolumn{3}{c}{Recon. + Animation} \\
            \cmidrule(lr){2-4} \cmidrule(lr){5-7}
            & \multicolumn{3}{c}{TH2.1 $\rightarrow$ TH4.0} & \multicolumn{3}{c}{TH2.1 $\rightarrow$ TH4.0} \\
            Methods & PSNR$\uparrow$ & SSIM$\uparrow$ & LPIPS$\downarrow$ & PSNR$\uparrow$ & SSIM$\uparrow$ & LPIPS$\downarrow$ \\
            \midrule
            HumanRAM & 28.98 & 0.97 & 0.021 & 25.10 & 0.96 & 0.029 \\
            LVSM     & 24.38 & 0.96 & 0.036 & -     & -    & -     \\
            GHG      & 22.49 & 0.95 & 0.043 & -     & -    & -     \\
            HumanGS  & 27.99 & 0.97 & 0.025 & 25.53 & 0.96 & 0.029 \\
            \bottomrule
        \end{tabular}}
        \caption{\textbf{Cross-dataset Evaluation.} Evaluated on TH4.0, HumanGS demonstrates highly competitive zero-shot generalization.}
        \label{tab:quantitative_results-th4}
    \end{minipage}
    \hfill
    \begin{minipage}[b]{0.48\textwidth}
    \centering
    \resizebox{\linewidth}{!}{
    \begin{tabular}{lcccccc}
        \toprule
        & \multicolumn{3}{c}{Reconstruction (TH2.1 $\rightarrow$ TH2.1)} & \multicolumn{3}{c}{Reconstruction (TH2.1 $\rightarrow$ AvRex)} \\
        \cmidrule(lr){2-4} \cmidrule(lr){5-7}
        Methods & PSNR$\uparrow$ & SSIM$\uparrow$ & LPIPS$\downarrow$ & PSNR$\uparrow$ & SSIM$\uparrow$ & LPIPS$\downarrow$ \\
        \midrule
        HumanRAM & 33.16 & 0.98 & 0.018 & 27.76 & 0.97 & 0.029 \\
        LVSM     & 27.13 & 0.95 & 0.051 & 20.86 & 0.94 & 0.056 \\
        GHG      & 26.91 & 0.96 & 0.039 & 22.69 & 0.93 & 0.088 \\
        GHG$^*$  & 22.80 & 0.91 & 0.067 & -     & -    & -     \\
        GST$^*$  & 21.58 & 0.92 & 0.091 & -     & -    & -     \\
        HumanGS  & 30.81 & 0.98 & 0.024 & 27.27 & 0.97 & 0.033 \\
        \midrule
        GIGA$^{**}$ & 20.19 & 0.794 & 0.070 & - & - & - \\
        HumanGS$^{**}$ & 26.00 & 0.945 & 0.061 & - & - & - \\
        \bottomrule
    \end{tabular}}
    \caption{\textbf{In-domain and Cross-domain Reconstruction.} HumanGS significantly outperforms static and single-image baselines (GIGA$^{**}$ is evaluated on its own train-test split; HumanGS$^{**}$ is retrained on this split for fair comparison).}
    \label{tab:quantitative_results-reco}
\end{minipage}
\end{table*}

\subsection{Reconstruction Evaluation} 
We evaluate reconstruction performance under the fixed-pose novel view synthesis (NVS) setting, where inputs are multi-view images of a single pose and targets are unseen viewpoints. We compare HumanGS against HumanRAM, GHG, GIGA, GST and LVSM using standard NVS metrics (PSNR, SSIM, LPIPS). Evaluation spans in-domain reconstruction (THuman 2.1 $\rightarrow$ THuman 2.1 in~\tabref{tab:quantitative_results-reco}) and cross-dataset generalization to THuman 4.0 (~\tabref{tab:quantitative_results-th4}) and AvatarReX (~\tabref{tab:quantitative_results-reco}). For completeness, we also report results using the publicly released pretrained GHG model (denoted as GHG$^*$). 

Across all datasets, HumanGS achieves performance comparable to HumanRAM and consistently outperforms the feed-forward 3DGS baseline GHG (~\figref{fig:qualitative_comparison}). Mirroring observations from the animation setting, our method maintains strong reconstruction fidelity while preserving the efficiency advantages of a feed-forward canonical 3D representation.

\subsection{Ablation Study}

To rigorously evaluate the architectural contributions of HumanGS, we conduct ablation studies focusing on two primary components: the integration of a global context token and the aggregation of intermediate network features. 





\subsubsection{Aggregation of Intermediate Features.}
Relying solely on a vision transformer's final layer can over-smooth high-frequency details critical for synthesizing complex clothing textures. Inspired by HumanRAM, our decoder aggregates intermediate tokens from multiple depths. Fusing and upsampling these multi-scale features creates a dense, high-resolution spatial map for sampling pixel-aligned vertex features, preserving fine-grained details otherwise bottlenecked by terminal representations.

\begin{figure}[h]
\centering
\includegraphics[width=\columnwidth]{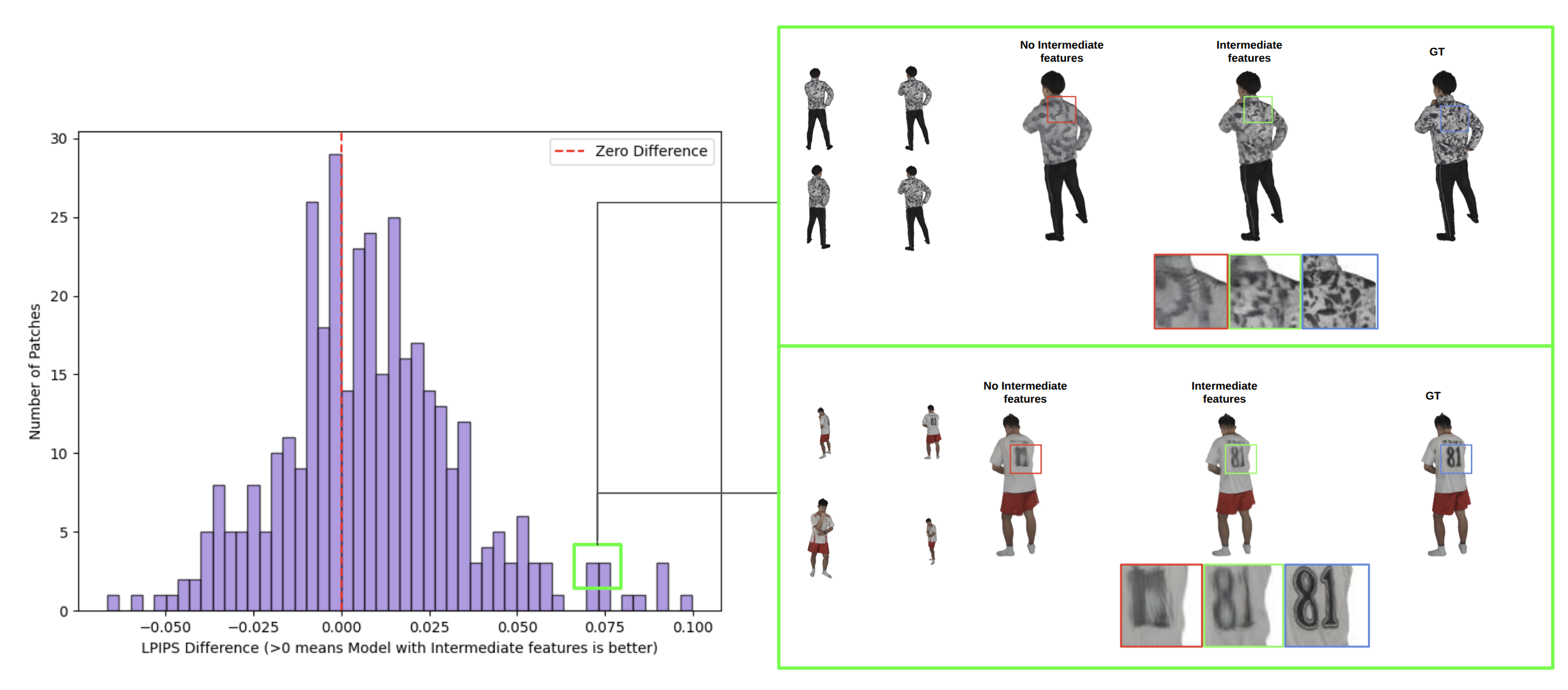}
\caption{Ablation of Intermediate Feature Aggregation. \textbf{Left:} Distribution of localized LPIPS differences on $64 \times 64$ foreground patches (positive difference indicates superior quality). \textbf{Right:} Qualitative comparisons from the distribution's high-performance tail. Intermediate features successfully recover high-frequency textures blurred by the baseline.}
\label{fig:intermediate_ablation}
\end{figure}

To rigorously evaluate this, we introduce a localized patch-based metric, as global metrics often obscure local textural gains. We divide rendered outputs into $64 \times 64$ patches, isolate those with $>50\%$ ground-truth foreground overlap, and compute the LPIPS difference between our full model and a baseline lacking intermediate aggregation. As shown in Figure~\ref{fig:intermediate_ablation}, the localized LPIPS differences exhibit a definitive positive skew (left), confirming enhanced perceptual fidelity. Qualitatively (right), while the baseline struggles with intricate textures---producing blurry high-frequency patterns---integrating multi-scale features successfully restores the sharpness and structural integrity of these local details.

\vspace{-5mm}

%% file: sec/5_conclusion.tex
\section{Conclusion}
We presented HumanGS, a feed-forward framework for high-fidelity human reconstruction and real-time animation. Unlike HumanRAM's repeated per-frame inference, HumanGS predicts an explicit canonical 3D Gaussian representation, decoupling representation learning from rendering to enable highly efficient animation via Linear Blend Skinning and rasterization. Compared to optimization-based Animatable Gaussians, our approach reduces subject-specific modeling time from minutes to seconds. Furthermore, while recent foundational models like LHM achieve impressive visual fidelity, HumanGS maintains a significant computational advantage, accelerating the initial modeling process by nearly $15\times$.

Extensive evaluations on THuman 2.1, THuman 4.0, and AvatarReX show HumanGS achieves reconstruction and animation quality comparable to state-of-the-art feed-forward and optimization-based pipelines. Despite training solely on fixed-pose data, it generalizes effectively to unseen subjects, poses, and camera viewpoints across datasets. These results confirm explicit canonical Gaussians provide a scalable, practical solution for real-time avatar generation. Future work will explore improved temporal modeling for complex non-rigid deformations.

%% file: sec/supply.tex
\setcounter{page}{1}


\maketitle





\section{Experiment Details}

\noindent \textbf{Evaluation Setup.} In order to rigorously and comprehensively evaluate our proposed methodology, we conduct an extensive series of experiments across three distinct and diverse datasets. Specifically, for the THuman 2.1~\cite{thuman21} dataset, we carefully partition the available data into a split of 2,300 subjects dedicated to training and 145 subjects reserved for testing. Furthermore, to effectively evaluate the cross-dataset generalization capabilities and the overall fidelity of our animation framework, we utilize the entirety of both the THuman 4.0~\cite{thuman4} and AvatarReX~\cite{avatarrex} datasets exclusively for testing purposes. This evaluation set encompasses all 3 subjects from THuman 4.0 and all 4 subjects from AvatarReX. For our standard geometric reconstruction and animation experiments, we consistently operate under a highly sparse four-view ($N=4$) setting. To ensure a completely fair and consistent comparative analysis against all established baseline methods, the specific input context views and the target novel evaluation views are strictly fixed and held constant across all conducted experiments.

In order to facilitate the ablation study examining the specific impact of the global token across a varying number of input views ($N \in \{1, 2, 4, 6\}$), we introduce a purposely modified THuman 2.1 test set designed to aggressively stress-test the rendering fidelity of the model. To demand the synthesis of much finer, high-frequency details, the human subjects in this modified set are positioned at a significantly closer proximity to the cameras. Specifically, the camera-to-subject distance is rigidly fixed at exactly twice the bounding sphere radius of the subject, whereas our standard evaluation protocol allows this distance multiplier to be uniformly sampled between $1.0$ and $1.75$.

\vspace{3mm}

\noindent \textbf{Data Pre-processing.} As a preliminary step in our data pre-processing pipeline, the raw ground truth 3D scans are first semantically aligned to a consistent, standardized orientation utilizing the corresponding SMPL-X joint locations. Following this alignment, we procedurally render 60 multi-view images per individual subject at a fixed spatial resolution of $512 \times 512$. The camera viewpoints for these renderings are distributed by uniformly sampling the azimuth angle from $0^\circ$ to $360^\circ$ and the elevation angle from $-5^\circ$ to $15^\circ$, with the camera-to-subject distance dynamically scaled in accordance with the bounding sphere radius of the aligned mesh.

\noindent \textbf{Training Details.} All computational experiments and training procedures for the proposed model are executed leveraging a single 48GB RTX 6000 GPU to ensure hardware consistency. To accelerate the rate of convergence and efficiently manage GPU memory overhead, we strategically employ a progressive resolution training strategy. The neural network is initially pre-trained from scratch on the THuman 2.1 dataset at a reduced lower resolution of $256 \times 256$ for a total of $60,000$ iterations using a batch size of $8$, a process requiring approximately 32 hours of compute time. Subsequently, the model undergoes a fine-tuning phase at the final target resolution of $512 \times 512$ for an additional $10,000$ iterations with a proportionally reduced batch size of $4$. Throughout the entirety of the training process, the network parameters are optimized utilizing the AdamW optimizer with a base learning rate of $4 \times 10^{-4}$, combined with a cosine learning rate decay scheduler and a standard weight decay factor of $0.05$.

\section{Additional Results}

\subsection{Robustness to Varying Input Views}
\begin{figure*}[t]
\centering
\includegraphics[width=\textwidth]{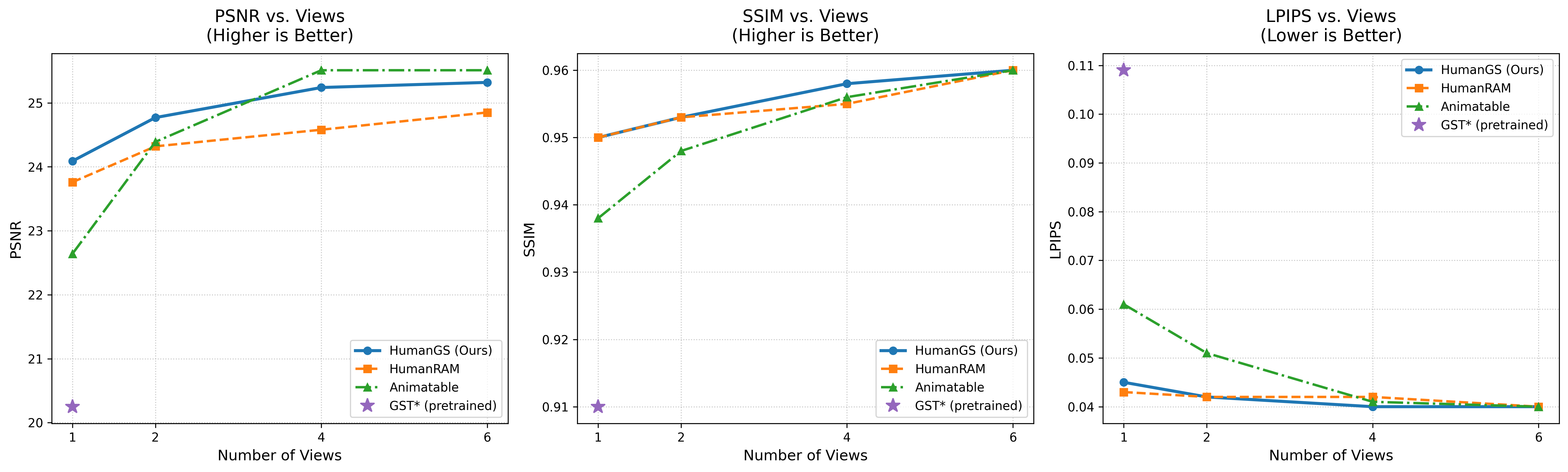}
\caption{\textbf{Generalization Performance Comparison Across Varying View Counts on AvatarReX (Animation).} We evaluate PSNR, SSIM, and LPIPS under 1, 2, 4, and 6 input views with varying poses on unseen subjects from the AvatarReX dataset. HumanGS consistently outperforms feed-forward baselines HumanRAM and GST. While the optimization-based Animatable Gaussians achieves comparable metrics given sufficient data (4 to 6 views), it suffers a catastrophic drop in performance under extreme sparsity (1 to 2 views), whereas our method remains robust. Note that the single-image baseline GST is reported only at $N=1$.}
\label{fig:varying_views}
\end{figure*}

In this section, we provide extended quantitative and qualitative results to further validate the robustness and generative capabilities of HumanGS. First, we evaluate the performance of our method under varying degrees of input sparsity ($N \in \{1, 2, 4, 6\}$). As shown in Figure~\ref{fig:varying_views}, when reconstructing subjects from the AvatarReX dataset with inputs featuring varying poses, HumanGS consistently outperforms the feed-forward baseline, HumanRAM~\cite{yu2025humanram}. Crucially, our method remains highly robust even in extreme 1- and 2-view scenarios where optimization-based methods like Animatable Gaussians~\cite{li2023animatable} suffer catastrophic degradation due to insufficient data. 

We further corroborate this robustness in Table~\ref{tab:thuman_near_views}, which details our stress-test evaluation on a modified THuman~2.1 test set where subjects are positioned closer to the cameras. Under these demanding conditions, HumanGS demonstrates stable, high-fidelity geometry and detail synthesis across all view counts compared to prior feed-forward approaches. We additionally compare against GST~\cite{prospero2024gst}, using its publicly available pretrained weights derived from its respective training dataset. As GST is architecturally restricted to a single input view, its performance is only reported at $N=1$. We used the official pre-trained weights of GST for this evaluation.

\begin{table*}[t]
\centering
\resizebox{\textwidth}{!}{
\begin{tabular}{l | ccc | ccc | ccc | ccc}
\hline
\multirow{2}{*}{Methods} & \multicolumn{3}{c|}{1 View} & \multicolumn{3}{c|}{2 Views} & \multicolumn{3}{c|}{4 Views} & \multicolumn{3}{c}{6 Views} \\
\cline{2-13}
 & PSNR $\uparrow$ & SSIM $\uparrow$ & LPIPS $\downarrow$ & PSNR $\uparrow$ & SSIM $\uparrow$ & LPIPS $\downarrow$ & PSNR $\uparrow$ & SSIM $\uparrow$ & LPIPS $\downarrow$ & PSNR $\uparrow$ & SSIM $\uparrow$ & LPIPS $\downarrow$ \\
\hline
HumanRAM       & 25.44 & 0.946 & 0.052 & 27.39 & 0.950 & 0.044 & 29.10 & 0.960 & 0.039 & 29.90 & 0.966 & 0.036 \\
LVSM           & 17.78 & 0.900 & 0.101 & 22.64 & 0.930 & 0.066 & 25.96 & 0.950 & 0.055 & 27.03 & 0.950 & 0.051 \\
GHG            & 21.71 & 0.920 & 0.094 & 22.43 & 0.930 & 0.089 & 22.69 & 0.930 & 0.088 & 22.72 & 0.930 & 0.088 \\
HumanGS (Ours) & 24.07 & 0.936 & 0.068 & 25.71 & 0.946 & 0.059 & 27.19 & 0.950 & 0.054 & 27.84 & 0.950 & 0.051 \\
\hline
\end{tabular}
}
\vspace{2mm}
\caption{\textbf{Reconstruction Performance Across Varying View Counts.} Quantitative comparison across varying numbers of input views ($N \in \{1, 2, 4, 6\}$) on the modified THuman 2.1 test set, where subjects are positioned closer to the cameras to evaluate fine-grained detail synthesis.}
\label{tab:thuman_near_views}
\end{table*}

\begin{figure*}[t]
\centering
\includegraphics[width=\linewidth]{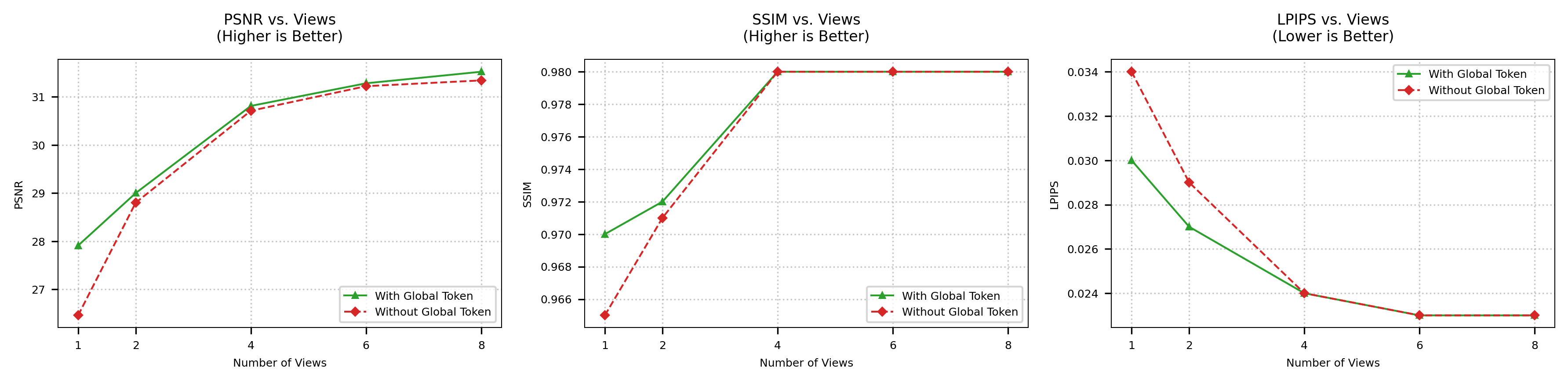}
\caption{\textbf{Effectiveness of the Global Token Under Varying Input Sparsity.} We evaluate reconstruction performance with and without the Global Token across 1 to 8 input views. By encoding holistic subject context, the token provides a crucial fallback for severely occluded regions in extreme sparse-view scenarios ($N \in \{1, 2\}$), yielding significant improvements. As spatial coverage increases ($N \ge 4$) and occlusions naturally decrease, the performance margins converge.}
\label{fig:views_ablation}
\end{figure*}


\subsection{Impact of the Global Token}
Acquiring local vertex features via 2D projection suffers a fundamental limitation under sparse-view conditions (e.g., 1 or 2 input views), where severe self-occlusion prevents a significant subset of vertices from sampling meaningful features. To prevent the resulting degenerate geometry and rendering artifacts, we concatenate a global token---encoding holistic subject context---to the local features, which acts as a robust fallback for these occluded regions.

We evaluated models trained with and without the global token across varying input views ($N \in \{1, 2, 4, 6, 8\}$) using a modified THuman 2.1 test set with closer camera positioning to stress-test fidelity. As shown in Figure~\ref{fig:views_ablation}, the performance gap is most pronounced in highly sparse 1- and 2-view scenarios, where the global token yields substantial improvements across all metrics. At 4 and 6 views, increased spatial coverage naturally reduces occlusions and the network's reliance on global context, leading to converging performance margins.

\subsection{Ablation on Gaussian Primitives per Vertex}

\begin{table*}[t]
\centering
\small
\begin{tabular}{l | ccc | ccc}
\hline
\multirow{2}{*}{Dataset} & \multicolumn{3}{c}{1 Gaussian per Vertex} & \multicolumn{3}{c}{1 Tight + 4 Free Gaussians (Ours)} \\
\cline{2-7}
 & PSNR $\uparrow$ & SSIM $\uparrow$ & LPIPS $\downarrow$ & PSNR $\uparrow$ & SSIM $\uparrow$ & LPIPS $\downarrow$ \\
\hline
THuman 2.1 & 30.66 & 0.970 & 0.028 & \textbf{30.81} & \textbf{0.980} & \textbf{0.024} \\
THuman 2.1 (Near) & 27.09 & 0.950 & 0.062 & \textbf{27.19} & \textbf{0.950} & \textbf{0.054} \\
AvatarReX & 25.20 & 0.960 & 0.044 & \textbf{25.24} & \textbf{0.960} & \textbf{0.040} \\
THuman 4.0 & 25.42 & 0.960 & 0.034 & \textbf{25.53} & \textbf{0.960} & \textbf{0.029} \\
\hline
\end{tabular}
\vspace{5mm}
\caption{\textbf{Ablation on Gaussian Primitives per Vertex.} Comparing reconstruction and animation performance using a single Gaussian versus our proposed hybrid representation (1 tight + 4 free Gaussians). The hybrid approach consistently improves perceptual quality (lower LPIPS) across all datasets by better capturing loose clothing and structures that deviate from the parametric mesh.}
\label{tab:gaussian_ablation}
\end{table*}

Additionally, we present an extended quantitative ablation on the number of Gaussian primitives assigned per SMPL-X vertex in Table~\ref{tab:gaussian_ablation}. Comparing a naive single-Gaussian baseline to our proposed hybrid representation (1 tight + 4 free Gaussians), we observe consistent improvements across all four evaluation settings (THuman~2.1, THuman~2.1 Near, AvatarReX, and THuman~4.0). The hybrid formulation yields particularly noticeable gains in perceptual quality (indicated by lower LPIPS scores), validating our hypothesis that the unconstrained ``free'' Gaussians effectively capture loose clothing, hair, and other topological deviations from the underlying parametric body model, while the tightly regularized Gaussian maintains structural stability.

\subsection{Extended Qualitative Visualizations}
\begin{figure*}[p]
    \centering
    \begin{subfigure}{0.85\textwidth}
        \centering
        \includegraphics[width=\linewidth]{figs/reconstruction/reconstruction_2.png}
    \end{subfigure}
    \\ \vspace{0.2cm} 
    \begin{subfigure}{0.85\textwidth}
        \centering
        \includegraphics[width=\linewidth]{figs/reconstruction/reconstruction_5.png}
    \end{subfigure}
    \\ \vspace{0.2cm}
    \begin{subfigure}{0.85\textwidth}
        \centering
        \includegraphics[width=\linewidth]{figs/reconstruction/reconstruction_1.png}
    \end{subfigure}
    \\ \vspace{0.2cm}
    \begin{subfigure}{0.85\textwidth}
        \centering
        \includegraphics[width=\linewidth]{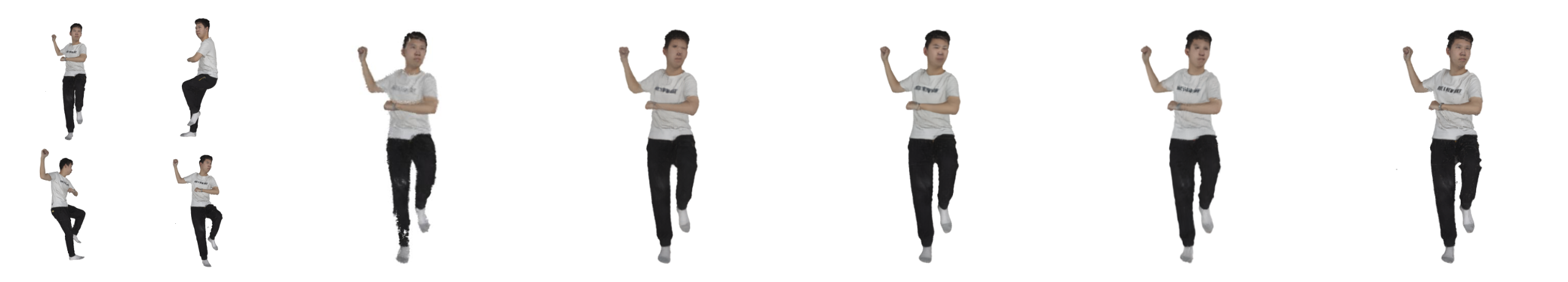}
    \end{subfigure}
    \begin{subfigure}{0.85\textwidth}
        \centering
        \includegraphics[width=\linewidth]{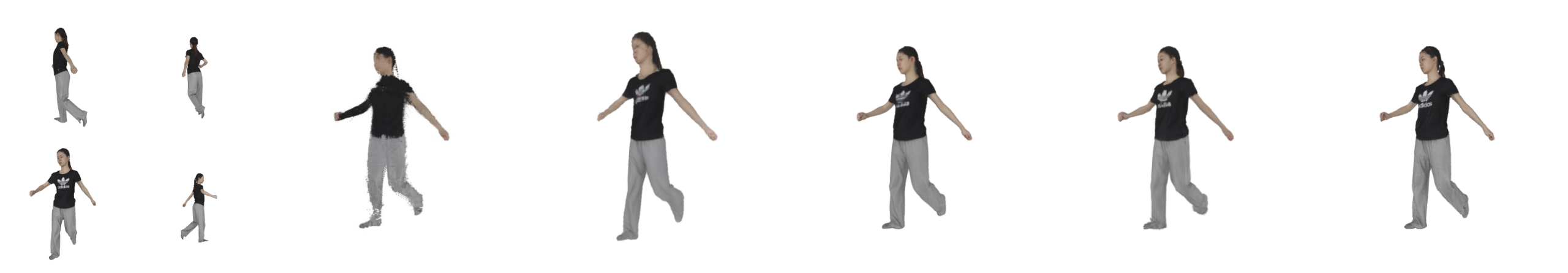}
    \end{subfigure}
    \\ \vspace{0.2cm} 
    \begin{subfigure}{0.85\textwidth}
        \centering
        \includegraphics[width=\linewidth]{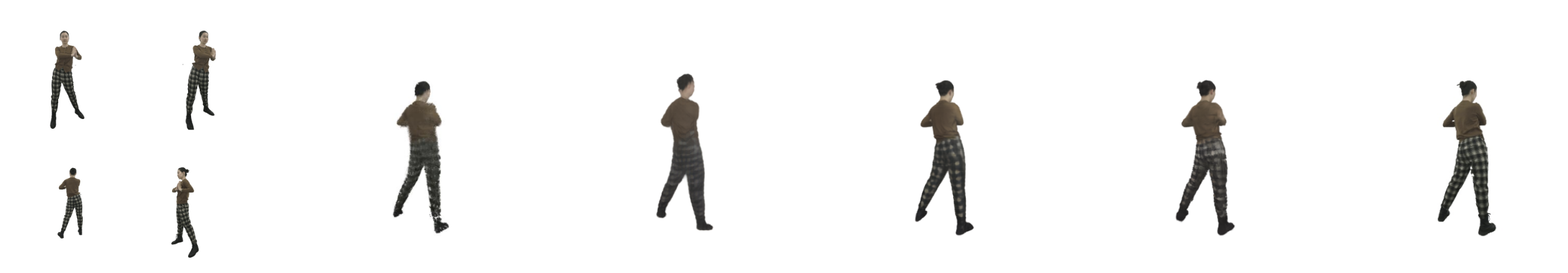}
    \end{subfigure}
    \\ \vspace{0.2cm}
    \begin{subfigure}{0.85\textwidth}
        \centering
        \includegraphics[width=\linewidth]{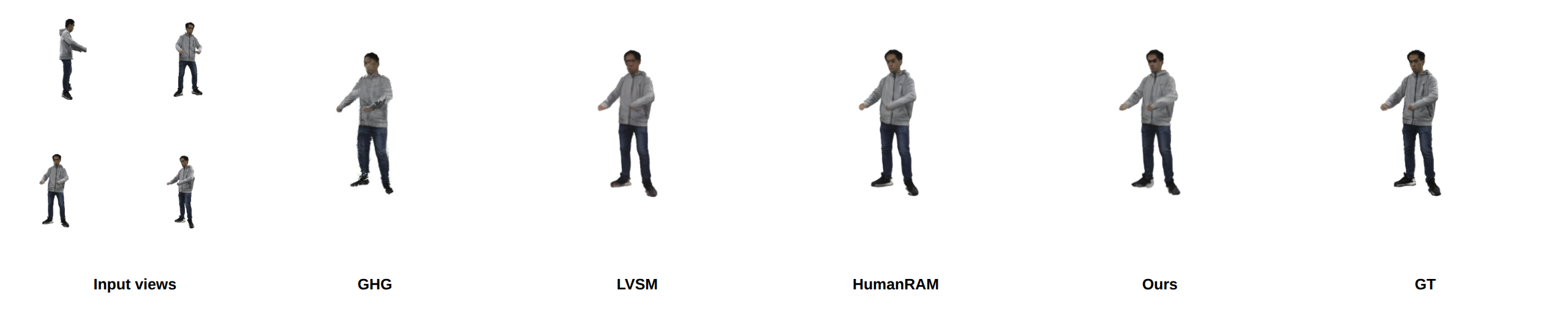}
    \end{subfigure}
    \caption{\textbf{Qualitative comparison on THuman2.1, AvatarReX, and THuman4.0.} For each subject, the left block shows the sparse input views. The right block displays novel view synthesis results from different methods. Our method achieves comparable visual quality to HumanRAM and superior visual quality to other methods while enabling real-time animation capabilities. The first four rows are from THuman2.1 and the last two rows are from AvatarReX and THuman4.0 respectively.}
    \label{fig:qualitative_comparison}
\end{figure*}

Finally, we provide comprehensive qualitative visualizations demonstrating both reconstruction and animation fidelity across diverse subjects and datasets. Figure~\ref{fig:qualitative_comparison} illustrates novel view synthesis comparisons across THuman~2.1, AvatarReX, and THuman~4.0, where HumanGS achieves visual quality comparable to or exceeding state-of-the-art baselines while uniquely preserving real-time animatability. 

\begin{figure*}[t]
    \centering
    \begin{subfigure}{0.88\textwidth}
        \centering
        \includegraphics[width=\linewidth]{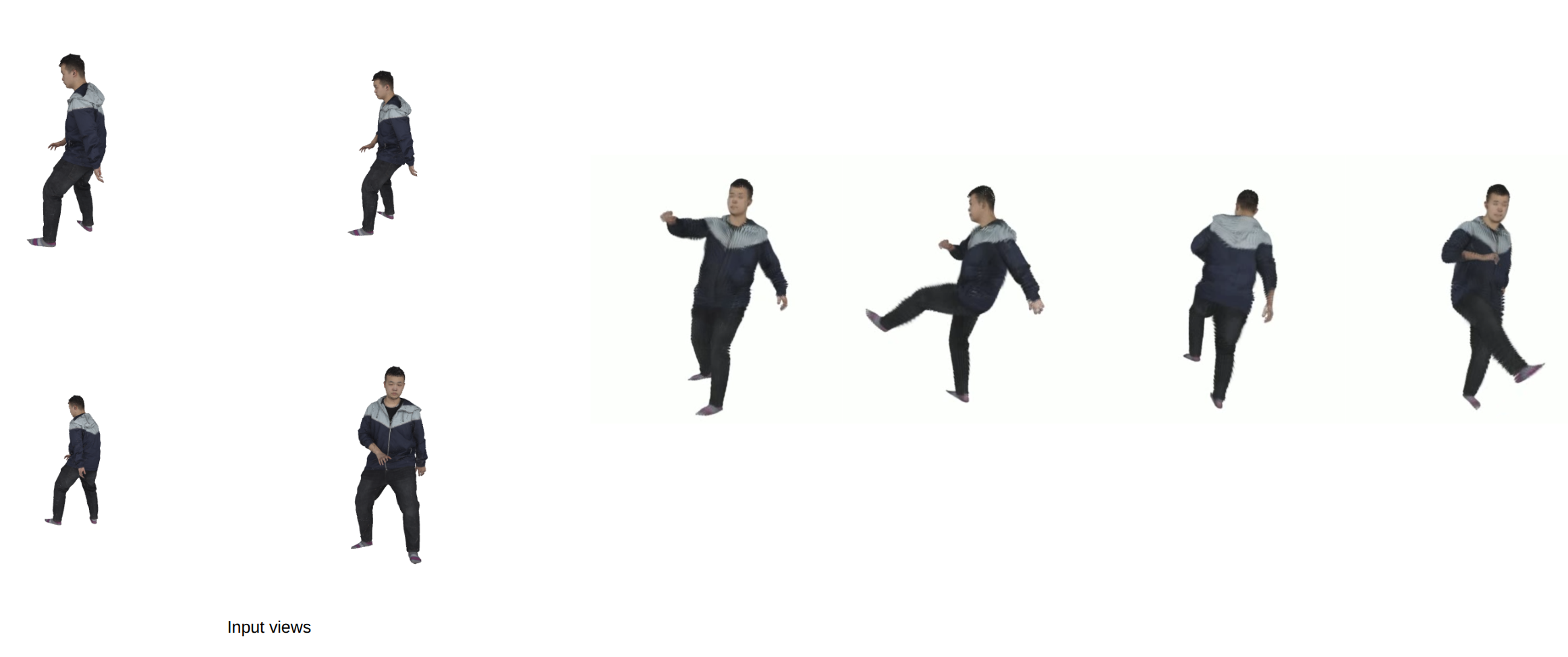}
    \end{subfigure}
    \\ \vspace{0.2cm}
    \begin{subfigure}{0.88\textwidth}
        \centering
        \includegraphics[width=\linewidth]{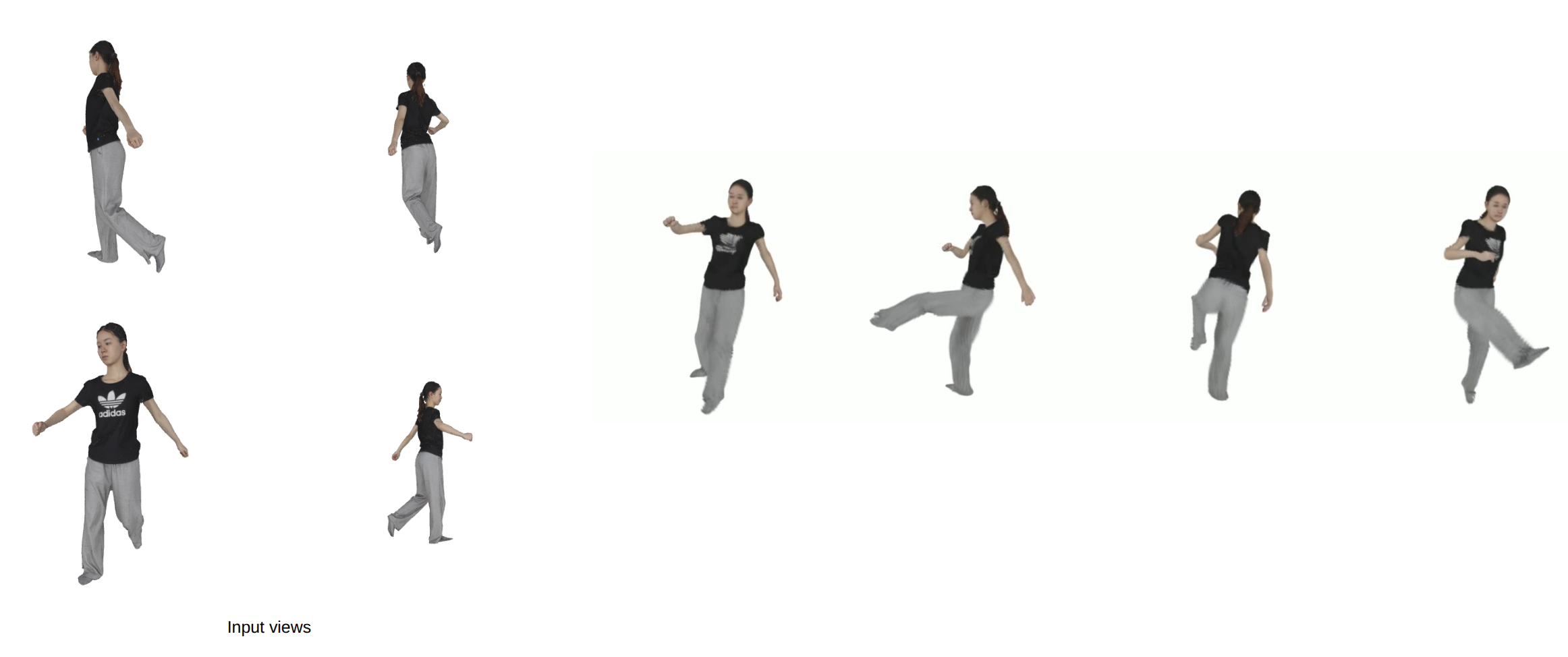}
    \end{subfigure}
    \\ \vspace{0.2cm}
    \begin{subfigure}{0.88\textwidth}
        \centering
        \includegraphics[width=\linewidth]{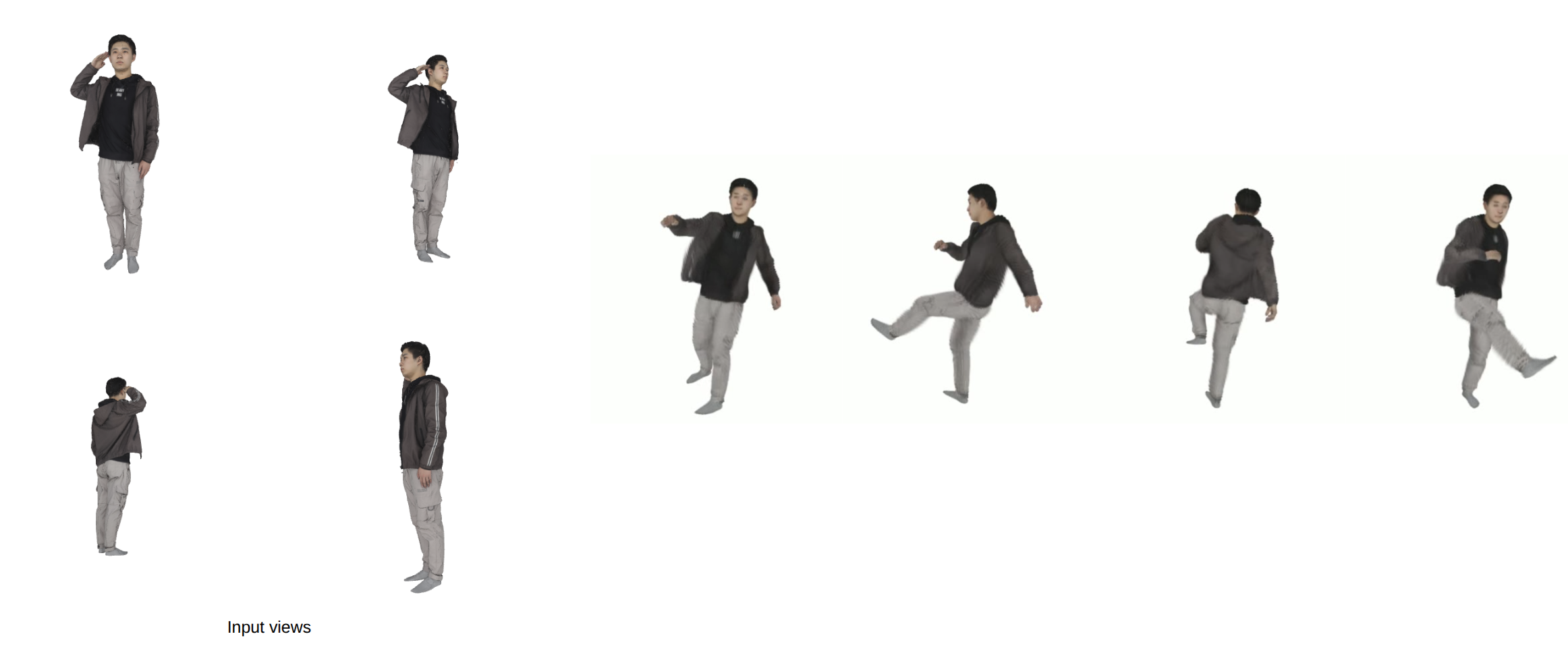}
    \end{subfigure}
    \caption{\textbf{Novel Pose Synthesis Results.} We demonstrate the explicit animation capabilities of HumanGS using subjects from THuman2.1 driven by motion sequences from THuman4.0. \textbf{Left:} The sparse set of 4 input views used to predict the canonical asset. \textbf{Right:} The predicted asset re-posed into various unseen configurations using Linear Blend Skinning and rasterization. This confirms that our method learns a consistent, animatable 3D representation from sparse supervision.}
    \label{fig:novel_pose_synthesis}
\end{figure*}

We further highlight this explicit animation capability in Figure~\ref{fig:avatarex_qualitative}, showcasing high-fidelity re-enactment on unseen AvatarReX subjects. Moreover, Figure~\ref{fig:novel_pose_synthesis} demonstrates successful novel pose synthesis, where static canonical assets predicted from sparse THuman~2.1 inputs are seamlessly driven by complex, out-of-distribution motion sequences extracted from THuman~4.0. These results collectively confirm that HumanGS learns a versatile, artifact-free 3D representation capable of robust cross-dataset generalization and highly expressive animation.

\subsection{Comparison with LHM}
\begin{figure*}[p]
    \centering
    \begin{subfigure}{0.58\textwidth}
        \centering
        \includegraphics[width=\linewidth]{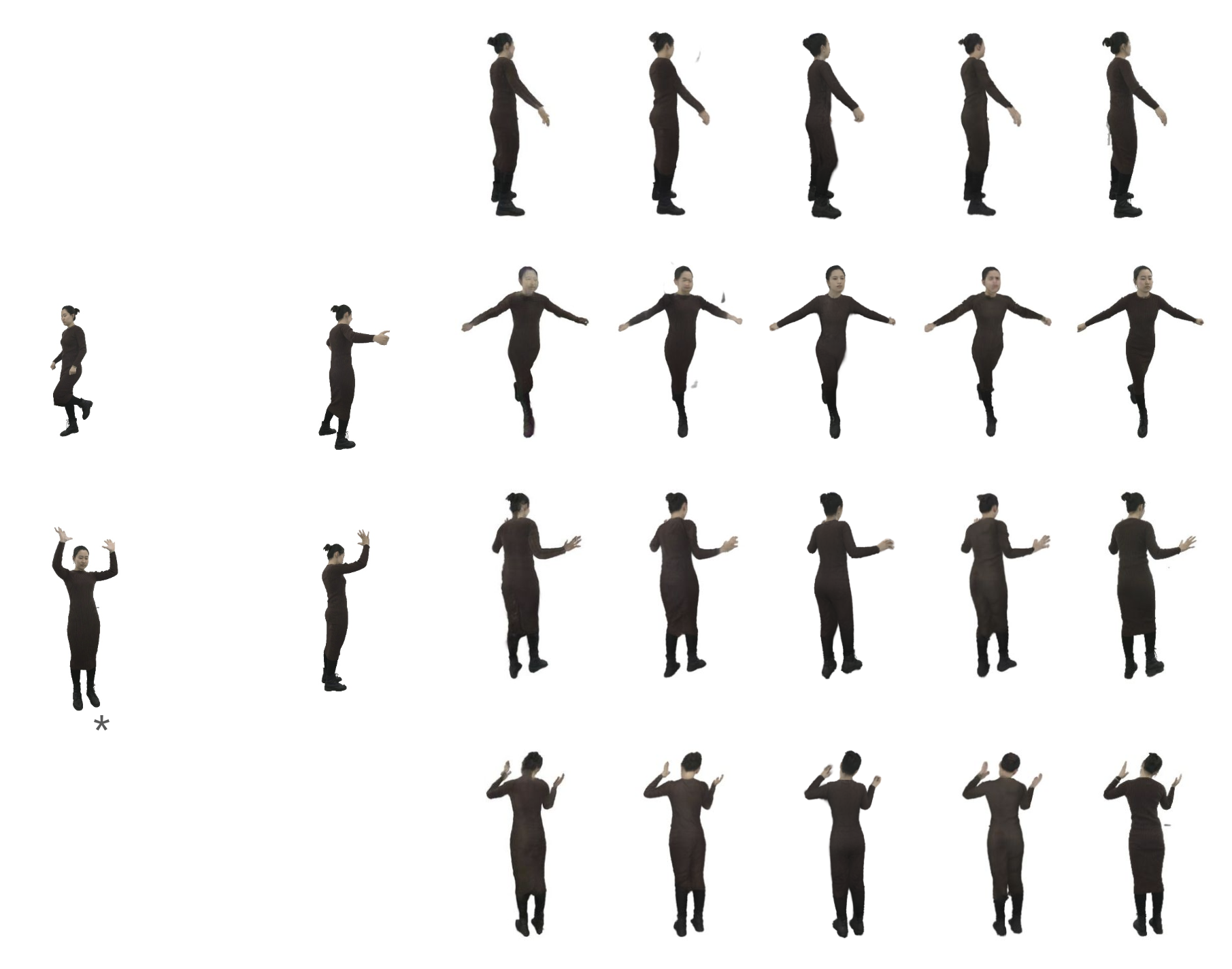}
    \end{subfigure}
    \\ \vspace{0.04cm}
    \begin{subfigure}{0.58\textwidth}
        \centering
        \includegraphics[width=\linewidth]{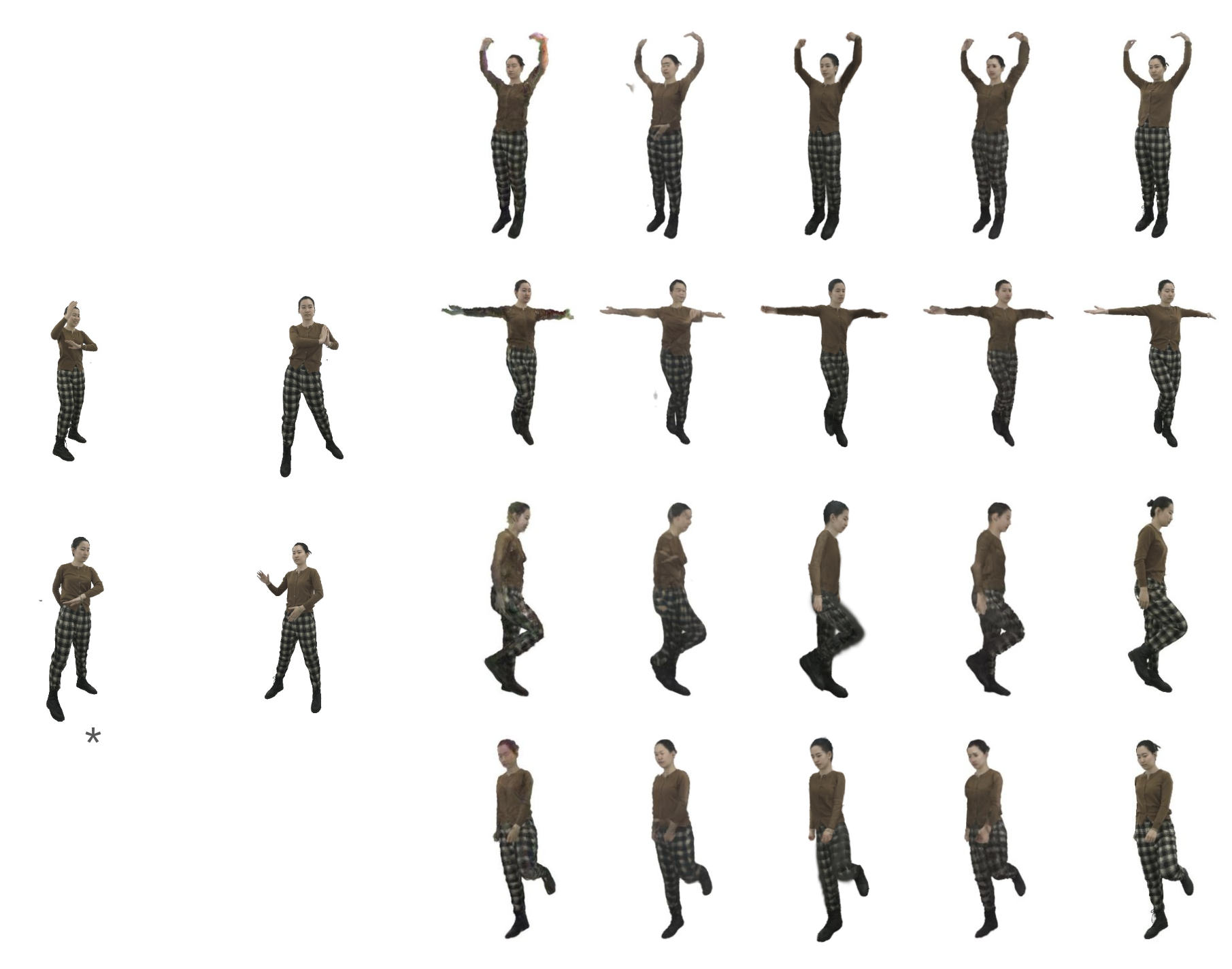}
    \end{subfigure}
    \\ \vspace{0.04cm}
    \begin{subfigure}{0.58\textwidth}
        \centering
        \includegraphics[width=\linewidth]{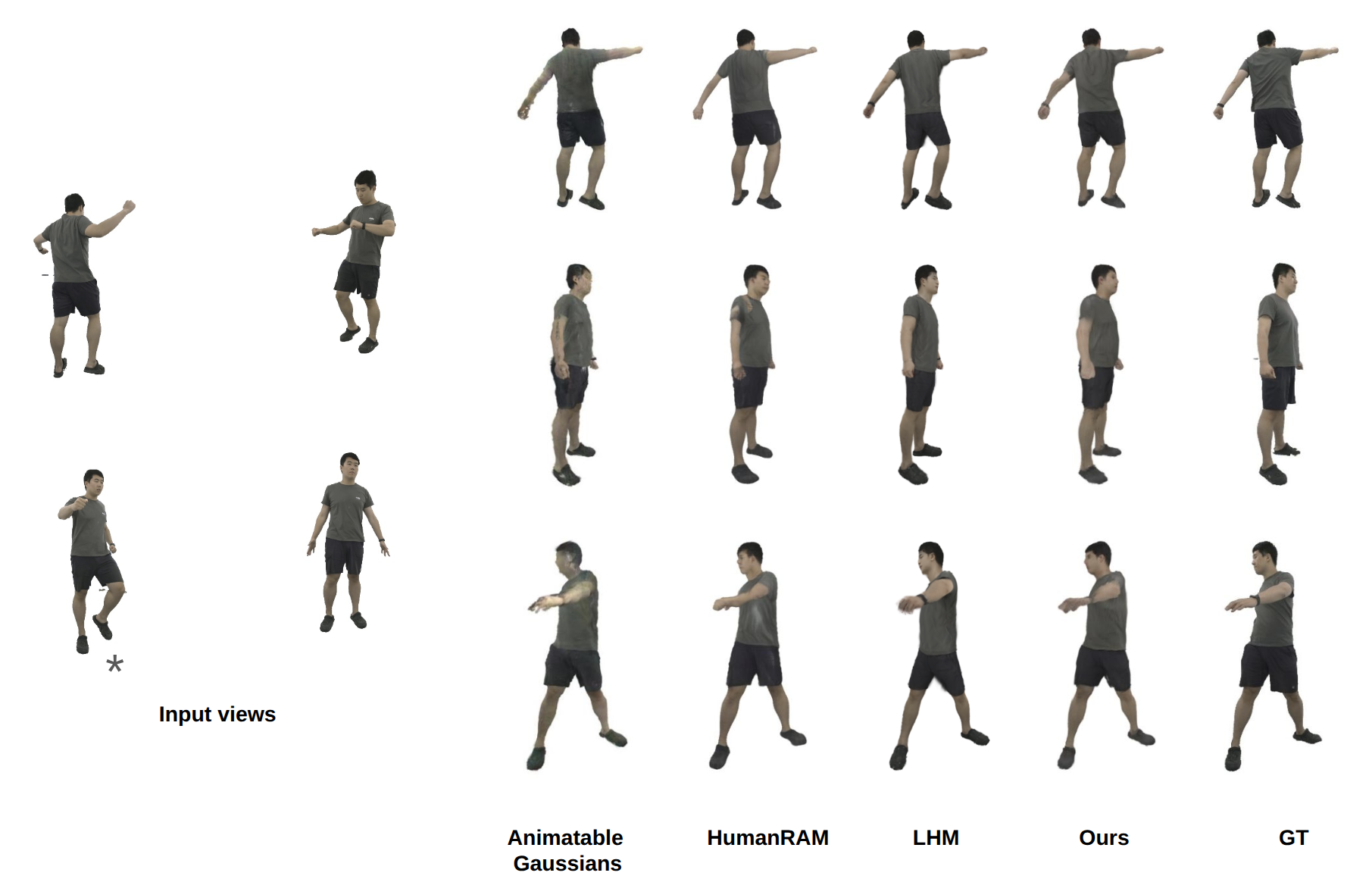}
    \end{subfigure}

    \caption{\textbf{Qualitative Generalization on AvatarReX.} \textbf{Left:} Input reference view. \textbf{Right:} Re-enactment results driven by a novel motion sequence.}
    \label{fig:avatarex_qualitative}
\end{figure*}

In addition to the visual multi-model comparisons provided in Figure~\ref{fig:avatarex_qualitative}, it is critical to properly contextualize our evaluation against the state-of-the-art single-image model, LHM~\cite{qiu2025lhm}, by considering both model capacity and overall training scale. 

For our experiments, we utilize the official pre-trained LHM-500M variant. This model was trained on an exceptionally massive dataset comprising approximately 300,000 videos and contains well over 500 million parameters in its core network alone---a figure that explicitly excludes the additional, computationally heavy Dinov2~\cite{oquab2023dinov2} and Sapiens~\cite{sapiens} backbones that it strictly requires for prior feature extraction. In stark contrast, our proposed HumanGS architecture operates with a significantly leaner total capacity of roughly 250 million parameters and is trained entirely from scratch on a mere 10,000 static frames. Furthermore, to ensure the fairest possible evaluation setting against a single-image baseline model, we explicitly restrict LHM's input to strictly front-facing camera views, guaranteeing that the competing model is consistently provided with the most informative and completely unoccluded optimal vantage point.

While standard global metrics successfully provide a macro-level evaluation of performance, visual results explicitly demonstrate our proposed method's robust preservation of high-frequency spatial details---such as intricate graphic prints, dynamic fabric wrinkles, and rigid facial boundaries---even during highly complex articulation sequences. Finally, this comparison highlights the practical runtime efficiency of HumanGS, which successfully achieves this elevated perceptual fidelity with a 15$\times$ faster modeling time (clocking in at 0.29s compared to LHM's 4.59s) by elegantly leveraging the canonical SMPL-X prior rather than relying on a cascade of computationally expensive feature extraction backbones.

\clearpage

